\documentclass[10pt,twocolumn,letterpaper]{article}

\usepackage[pagenumbers]{iccv} 

\newcommand{\name} {UnZipLoRA}
\newcommand{\SP} {\textless$s$\textgreater~}
\newcommand{\CP} {\textless$c$\textgreater~}
\newcommand{\R}{\mathbb{R}}

\usepackage{makecell}

\definecolor{iccvblue}{rgb}{0.21,0.49,0.74}
\usepackage[pagebackref,breaklinks,colorlinks,allcolors=iccvblue]{hyperref}
\newcommand{\nocontentsline}[3]{}
\let\origcontentsline\addcontentsline
\newcommand\stoptoc{\let\addcontentsline\nocontentsline}
\newcommand\resumetoc{\let\addcontentsline\origcontentsline}
\renewcommand{\thefootnote}{\arabic{footnote}}
\def\paperID{6056} 
\def\confName{ICCV}
\def\confYear{2025}

\title{UnZipLoRA: Separating Content and Style from a Single Image}

\author{Chang Liu, Viraj Shah\textsuperscript{*}, Aiyu Cui\textsuperscript{\textdagger}, Svetlana Lazebnik\\
University of Illinois, Urbana-Champaign\\
{\tt\small changl25, vjshah3, aiyucui2, slazebni@illinois.edu}
}

\begin{document}
\stoptoc 
\twocolumn[{%
\renewcommand\twocolumn[1][]{#1}%
\maketitle
\begin{center}
    \centering
    \vspace{-5mm}
    \includegraphics[width=0.99\textwidth]{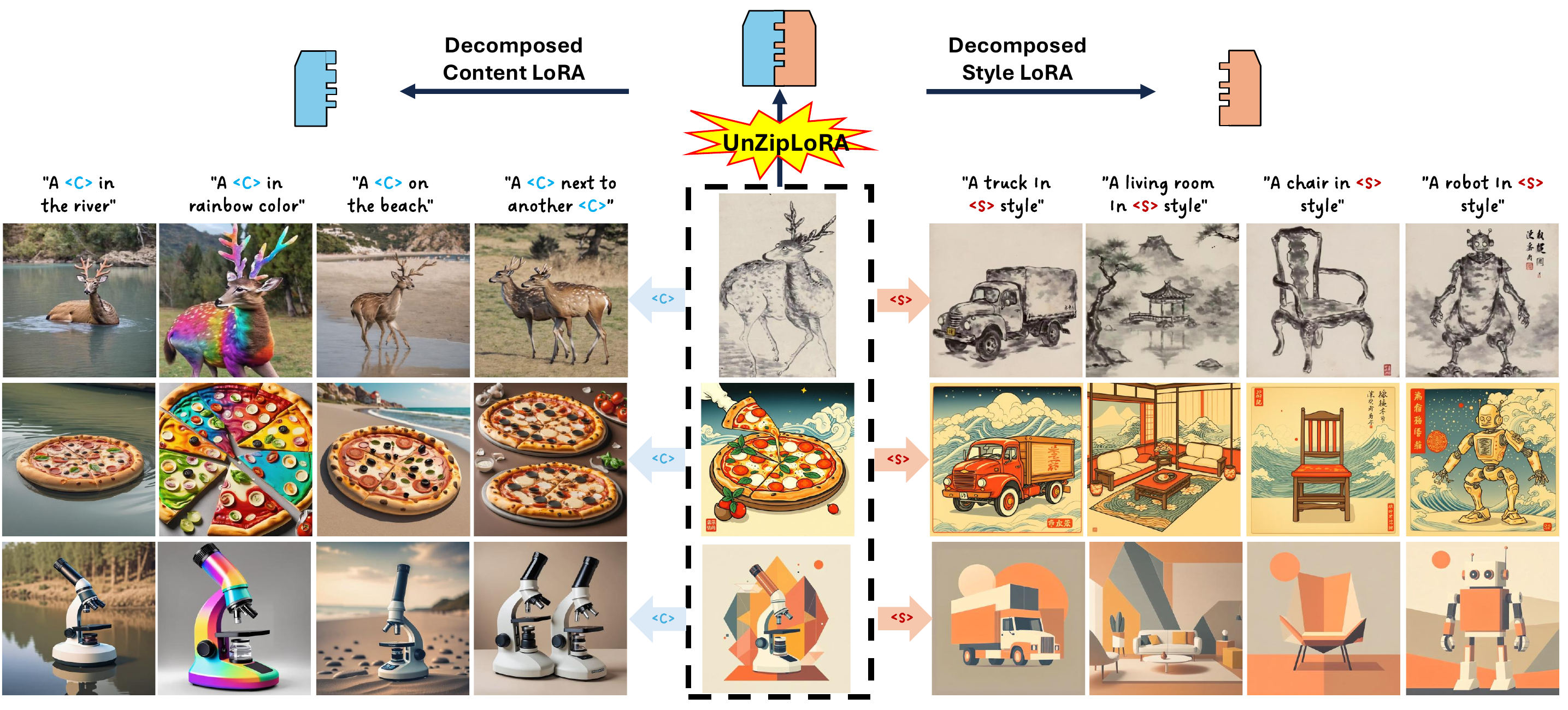}
    \vspace{-2mm}
    \captionof{figure}{Given a single input image (middle), \textbf{UnZipLoRA} learns a disentangled {\em content} or {\em subject} LoRA (left) and {\em style} LoRA (right) that can be used to generate new images with the learned concepts.}
    \label{fig:teaser}
\end{center}%
}]

\begin{abstract}
\let\thefootnote\relax\footnote{
\textsuperscript{*} Currently at Google. \quad \textsuperscript{\textdagger} Currently at Amazon. \\ \indent ~~~Project page: \url{https://unziplora.github.io} }
\footnote[2]{}
This paper introduces UnZipLoRA, a method for decomposing an image into its constituent subject and style, represented as two distinct LoRAs (Low-Rank Adaptations). Unlike existing techniques that focus on either subject or style in isolation, or require separate training sets for each, UnZipLoRA disentangles these elements from a single image by training both the LoRAs simultaneously. UnZipLoRA ensures that the resulting LoRAs are compatible, \ie, can be seamlessly combined using direct addition. UnZipLoRA enables independent manipulation and recontextualization of subject and style -- generating variations of each, applying the extracted style to new subjects, and recombining them to reconstruct the original image or create novel variations. To address the challenge of subject and style entanglement, UnZipLoRA employs a novel prompt separation technique, as well as column and block separation strategies to accurately preserve the characteristics of subject and style and ensure compatibility between the learned LoRAs. Evaluation with human studies and quantitative metrics demonstrates UnZipLoRA's effectiveness compared to other state-of-the-art methods, including DreamBooth-LoRA, Inspiration Tree, and B-LoRA.\vspace{-1em}
\end{abstract}    
\section{Introduction}
\label{sec:intro}
Imagine an artist inspired by a single image – perhaps a microscope rendered in a unique illustration technique (Fig.~\ref{fig:teaser}). The image depicts both the {\em content} or {\em subject} (the microscope with its specific shape and details) and {\em style} (the characteristic artistic technique). What if we could disentangle these intertwined elements, enabling the artist to extract and manipulate them independently? This capability, long sought in the computer vision literature~\cite{tenenbaum1996style}, would open up new avenues for artistic expression, allowing for the destylization and recontextualization of the subject in different conditions, the application of the extracted style to new subjects, and the creation of entirely novel combinations while preserving both the original subject and style. 

The recent surge of diffusion models~\cite{Ho2020DenoisingDP, Song2020DenoisingDI, Rombach2021HighResolutionIS} has unlocked previously unprecedented automatic image generation capabilities. The primary means of controlling such models is through text prompts, but text-based conditioning is inadequate for capturing the details of nuanced concepts, such as specific object instances (my dog) or individual styles (my child's crayon drawing). Example-driven generation is highly desired in such scenarios. To this end, model fine-tuning methods like DreamBooth~\cite{ruiz2023dreambooth} and StyleDrop~\cite{sohn2023styledrop} capture subject or style from reference image(s) in a way that allows for novel renditions. However, such approaches tend to focus on either content or style in isolation and cannot be easily made to capture both or perform disentanglement. On the other hand, stylization techniques like ZipLoRA~\cite{shah2024ziplora} and B-LoRA~\cite{Frenkel2024ImplicitSS} can combine subject and style, but require the training of two separate models using subject- and style-specific input images. However, the scenario we consider requires the opposite -- decomposing the subject and style from a single image.

In this work, we introduce \textbf{UnZipLoRA}, a novel method that deconstructs an image into its constituent subject and style, represented as two distinct LoRAs (Low-Rank Adaptations~\cite{hu2022lora}) trained simultaneously. These LoRAs can be used independently to generate variations of the subject or style and allow for recontextualizations. Moreover, our joint training method ensures that the resulting LoRAs are inherently compatible, \ie, can be seamlessly combined by direct addition to reconstruct the original image or to generate novel compositions of subject and style while preserving their fidelity. In fact our approach can be seen as a next stage of \emph{concept extraction} -- a problem previously studied in concept decomposition methods like Inspiration Tree~\cite{Vinker2023ConceptDF} and CusConcept~\cite{Xu2024CusConceptCV} that rely on textual inversion~\cite{gal2022textual} to learn multiple text embeddings corresponding to the \emph{subconcepts} within a set of images. However, textual inversion alone, without fine-tuning of weights, does not provide adequate control over or fidelity of the extracted concepts, which tend to remain generic and fail to capture nuances of the input object/style. 

As suggested by the name, UnZipLoRA operates in the opposite direction of ZipLoRA~\cite{shah2024ziplora}, which is focused on merging independently trained subject and style LoRAs. While ZipLoRA addresses the challenge of combining pre-existing LoRAs, UnZipLoRA tackles the inverse problem: disentangling a single image into its subject and style components such that the resulting LoRAs can be used either together or separately. Mathematically speaking, the decomposition problem is ill-posed and cannot be trivially derived from the approach of ZipLoRA~\cite{shah2024ziplora}.


Our key challenge is to learn two independent LoRAs simultaneously using only a single input image as supervision while ensuring that the resulting LoRAs correctly capture the subject and style concepts. Typical LoRA fine-tuning operates by binding the LoRA weights with the trigger phrase representing a specific subject or style concept in the input image -- such as ``a \CP in \SP style". 
If we apply such a naive approach in our case, the presence of both the trigger phrases \CP and \SP in a single input prompt makes it difficult for the subject and style LoRAs to bind to the correct trigger phrase, resulting in cross-contamination. To solve this problem, we propose a novel \emph{prompt separation} strategy that uses different prompts for each LoRA and the base model, and then combines them together in the intermediate feature space of the diffusion model in such a way that the loss for each LoRA can be calculated jointly using only the input image as supervision. 

While prompt separation allows for joint training of LoRAs, the resulting LoRAs may not be compatible with each other, \ie, combining them through direct addition may produce poor quality recontextualizations. To make them inherently compatible, we also propose \emph{column separation} and \emph{block separation} strategies. In particular, 
{\em column separation} determines the importance of each column of LoRA weight matrix and adaptively assigns each column to either a subject LoRA or style LoRA using a dynamic importance re-calibration strategy. Such disjoint assignment ensures that high-importance columns from each LoRA remain decoupled. \emph{Block separation} reserves some blocks of the U-net predominantly for style or for the subject, providing a further degree of disentanglement. 

We demonstrate the effectiveness of UnZipLoRA for accurate separation of content and style using both human studies and quantitative metrics. Our results show a clear advantage of our method over separate LoRA fine-tuning via DreamBooth~\cite{ruiz2023dreambooth}, concept separation via Inspiration Tree~\cite{Vinker2023ConceptDF}, and even the most recent state-of-the-art B-LoRA method~\cite{Frenkel2024ImplicitSS}. We also showcase our method's ability to preserve the concept fidelity for a wide array of recontextualizations -- whether for using subject or style separately or together. In addition, our method provides the valuable capability for cross-composition of subject and style LoRAs obtained from different images. Finally, we demonstrate generalizability of our approach by providing results on KOALA -- a newer, more efficient text-to-image model~\cite{Lee@koala}. 








\section{Related Work}
\label{sec:relatedwork}

\noindent \textbf{Fine-tuning diffusion models} is an effective way to personalize the text-to-image (T2I) models to depict specific concepts based on textual descriptions. Textual Inversion~\cite{gal2022textual} optimizes the text embedding to represent a specific visual concept. DreamBooth~\cite{ruiz2023dreambooth} fine-tunes the diffusion model itself to better capture an input concept from a small number of images, while another group of methods~\cite{han2023svdiff,kumari2022customdiffusion} aim to optimize specific parts of the networks to capture visual concepts. Most personalization approaches have quickly adopted LoRA~\cite{hu2022lora}, a fine-tuning technique that only optimizes a small subset of weights by low-rank approximations, as it is efficient for training and can mitigate overfitting problems.

\noindent \textbf{Image stylization}
is an area of research dating back at least 20 years~\cite{efros2001image,Hertzmann2001ImageA}. Great advances in arbitrary style transfer were achieved by approaches based on convolutional neural networks~\cite{gatys2016image,johnson2016perceptual,huang2017arbitrary, WCT-NIPS-2017,park2019arbitrary}. With the advent of deep generative models, a variety of approaches have attempted to fine-tune a pre-trained Generative Adversarial Network (GAN) or diffusion model for stylization~\cite{Liu2021BlendGANIG, Gal2021StyleGANNADACD, zhu2022mind, Ojha2021FewshotIG, Wang2022CtlGANFA, Kwon2022OneShotAO,jojogan, multistylegan, pastiche, oneshotadaption, sohn2023styledrop, dong2023dreamartist}. While these works provide valuable insights into style learning using generative models, the task we attempt to solve is the opposite: instead of stylizing a content image, we attempt to decompose a stylized image into its subject and style.

\noindent \textbf{Content-style decomposition. }
The task of subject-style decomposition can also be seen as a type of \emph{concept extraction}. Inspiration Tree~\cite{Vinker2023ConceptDF} aims to learn multiple embeddings corresponding to hierarchical subconcepts within a set of images. However, its reliance on textual inversion~\cite{gal2022textual} limits this method to primarily producing text embeddings rather than the full LoRA weights needed for granular control over generation. 
Similarly, CusConcept~\cite{Xu2024CusConceptCV} decomposes an image into visual concepts by learning customized embeddings. However, it does not produce dedicated LoRAs for each concept. Its reliance on computationally expensive LLMs and lack of explicit content-style separation limits its practical applicability.
U-VAP~\cite{Wu2024UVAPUV} is a fine-tuning method that allows users to specify desired attributes from a set of images, enabling the disentangled use of visual concepts in diverse settings. However, this method requires elaborate data augmentation aided by LLMs, posing scalability challenges. 
ConceptExpress~\cite{hao2024conceptexpress} explores unsupervised concept extraction and recreation by leveraging the inherent capabilities of pre-trained diffusion models. However, its reliance on localized masks extracted from the U-Net limits its application to concepts that can be localized within an image. This constraint prevents it from effectively handling global concepts like style. This limitation applies to Break-A-Scene~\cite{Avrahami2023BreakASceneEM} too, which relies on localized masks.

The method most directly aimed at content and style separation is B-LoRA~\cite{Frenkel2024ImplicitSS}. The authors of B-LoRA analyze the architecture of the SDXL base model~\cite{Podell2023SDXLIL} to find U-Net blocks most responsible for capturing content and style. By \emph{independently} training content and style LoRAs restricted to the respective blocks, they obtain models that can be successfully mixed for various stylization applications. By contrast, UnZipLoRA trains both the LoRAs \emph{simultaneously}. In our attempt to take advantage of B-LoRA's block constraints for subject-style separation, we find that they are coarse-grained and thus insufficient to disentangle the subject in joint training. Therefore, we extend the block separation to more blocks and perform more fine-grained disentanglement. Together with our improved block separation strategy, the prompt separation strategy of UnZipLoRA provides a significant improvement in disentanglement ability, while column separation further enhances the fidelity of fine details.

\section{Method}
\label{sec:method}

\begin{figure*}[h]
    \centering
    \includegraphics[width=0.9\linewidth]{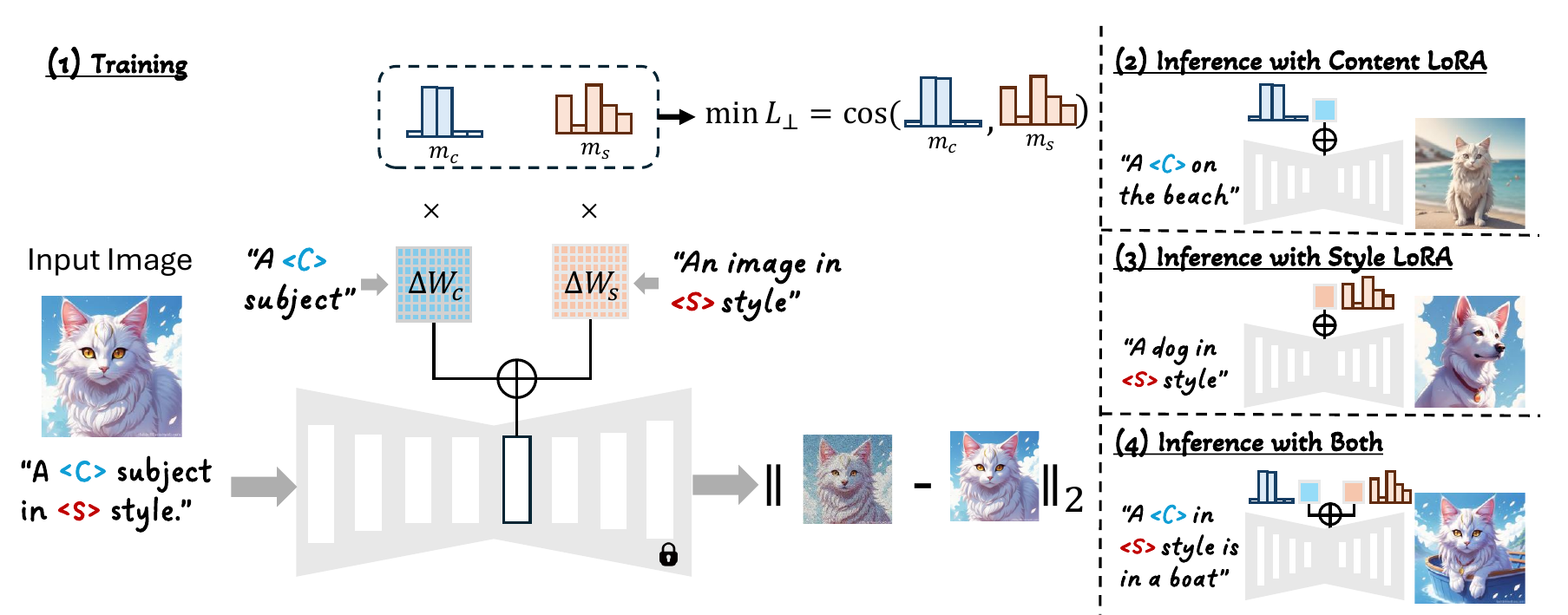}
    \caption{\textbf{Method Overview. }(1): Our training pipeline takes three prompts to learn two separate LoRAs. The column masks are introduced to establish the orthogonality. (2)-(4): Inference with individual learned LoRA or the combined LoRA.}
    \label{fig:method}
    \vspace{-1em}
\end{figure*}

\subsection{Preliminaries}
\noindent \textbf{Diffusion models}~\cite{Ho2020DenoisingDP, Song2020DenoisingDI, Rombach2021HighResolutionIS} are state-of-the-art deep generative models renowned for their ability to synthesize high-quality photorealistic images. 
In this work, we focus on Stable Diffusion XL (SDXL)~\cite{Podell2023SDXLIL}, a widely used U-Net-based latent diffusion model (LDM)~\cite{Rombach2021HighResolutionIS} known for its strong performance. In Section \ref{sec:extensions}, we also show results on the newer, more efficient KOALA model~\cite{Lee@koala}.

\noindent \textbf{Low-Rank Adaptation (LoRA)} is an efficient fine-tuning method for adapting large vision or language models to new tasks~\cite{hu2022lora,simo_lora}. LoRA assumes that weight updates $\Delta W$ during fine-tuning have a low intrinsic rank and can thus be decomposed into two low-rank matrices, $B \in \R^{m \times r}$ and $A\in \R^{r \times n}$, such that $\Delta W = BA$, where $r$ is the intrinsic rank of $\Delta W$ with $r << \min(m,n)$. During training, only $A$ and $B$ are updated while keeping the original weights $W_0$ constant. The updated weights become $W = W_0 + BA$. In case of text-to-image LDM models, model customization can be achieved by using LoRA fine-tuning to minimize reconstruction loss $\mathcal{L}_{DB}$ as proposed by DreamBooth~\cite{ruiz2023dreambooth}.

\subsection{Problem Setup}

We aim to extract content and style from a single input image by learning two distinct LoRAs simultaneously. Given a pre-trained diffusion model with weights $\{W_0^i\}$, we learn two models: content LoRA $L_c = \{\Delta W_c^i\}$ and style LoRA $L_s = \{\Delta W_s^i\}$. Here, $i$ denotes the index of layers of the diffusion U-Net. In the following, we will skip the superscript $i$ as we operate over all LoRA-enabled weights of the base model. Once trained, the resulting LoRAs can be used either separately or together to achieve various recontextualizations as depicted in Fig.~\ref{fig:method}.

We follow the standard prompt construction strategy ``a \CP in \SP style" with trigger phrases \CP and \SP to describe the content and style respectively~\cite{ruiz2023dreambooth,sohn2023styledrop,Frenkel2024ImplicitSS,kumari2022customdiffusion,shah2024ziplora}. Following prior works such as DreamBooth~\cite{ruiz2023dreambooth}, the subject trigger phrase \CP is formed with a unique token followed by the subject class label (e.g., `sks dog'), and the style trigger phrase \SP consists of a generic description of style such as `watercolor painting'~\cite{shah2024ziplora,sohn2023styledrop,Frenkel2024ImplicitSS}. We find that a single-word class label for the subject, and a high-level, brief (2-3 word) description for style are sufficient to effectively guide our method (see the supplementary material for further discussion of trigger phrase selection). 

\subsection{UnZipLoRA}
\label{sec:unziplora}
In this section, we present our UnZipLoRA approach, which relies on three key components to ensure accurate disentanglement. First, \textbf{prompt separation} (Sec.~\ref{sec:prompt-separation}) allows for training the subject and style LoRAs simultaneously. Second, \textbf{column separation} (Sec.~\ref{sec:weight-separation}) uses a dynamic importance recalibration strategy to make the columns of resulting LoRAs mutually orthogonal. Third, we introduce \textbf{block separation} (Sec.~\ref{sec:block-separation}) that reserves certain style/subject-sensitive blocks of the SDXL U-Net only for style/subject learning to improve the fine details.

\subsubsection{Prompt Separation}\label{sec:prompt-separation}
Our core challenge lies in the utilization of a single image as supervision for training two separate LoRAs. 
Since the input image contains both subject and style, 
we must use both LoRAs $L_c$ and $L_s$ together to calculate the loss during training. 
If we also use \CP and \SP in the input prompt, as in ``A \CP in \SP style", the training would lead to cross-contamination since the tokens corresponding to the style descriptor \SP would be attended to by the cross-attention layers of the subject LoRA, and vice versa. This is evident in the Baseline (DreamBooth-LoRA) row of Fig.~\ref{fig:ablation}, where both the subject and style LoRAs are overfitted to input image. 

Cross-attention layers in diffusion models are responsible for learning the text conditioning, and play a crucial role in binding the target concepts to the corresponding parts of the prompt. 
In a typical cross-attention layer of the diffusion U-Net, the prompt embedding $x$ is mapped to keys $K$ and values $V$ in the transformer using weights $W_0$:
\begin{equation}
    K(x)~\textnormal{or}~V(x) = W_0^T x \,.
\end{equation} 

If we add the content and style LoRAs to the base model and use both \CP and \SP in the prompt $x$, the mapping in the cross-attention layer becomes
\begin{equation}\label{eq:prompt_naive}
    K(x)~\textnormal{or}~V(x) = \left(W_0 + \Delta W_c + \Delta W_s\right)^T x \,.
\end{equation}
This allows the content (resp. style) LoRA to attend to style (resp. content) tokens, resulting in cross-contamination. 
Instead of the naive strategy in eq. (\ref{eq:prompt_naive}), we propose to calculate three sets of keys and values using three separate prompts: one with the base model $W_0$ and combined prompt $x$, and one each with the content and style LoRA and their respective prompts (See Fig.~\ref{fig:method}). The resulting feature maps are then added together as
\begin{equation}\label{eq:prompt_sep}
    K~\textnormal{or}~V(x, x_s, x_c) = W_0^T x +  \Delta W_s^T x_s +  \Delta W_c^T x_c,
\end{equation}
where  $x$ is the embedding for the combined prompt ``A \CP in \SP style", while $x_c$ and $x_s$ are the embeddings of subject and style descriptors \CP and \SP respectively. This allows each LoRA to attend to different concepts. As illustrated in Fig.~\ref{fig:ablation} (row M1), adding prompt separation to a DreamBooth baseline prevents cross-contamination and successfully destylizes the content. 

\subsubsection{Column Separation}\label{sec:weight-separation}
Apart from learning separate concepts, we want our resulting LoRAs to be compatible, \ie we want to be able to combine them through direct arithmetic merge to generate the subject and style together. 
Prompt separation effectively guides the two LoRAs to learn distinct concepts, and especially helps in achieving better destylization of the content. However, it does not guarantee compatibility since training LoRAs with different prompts can result in weight misalignment when they are combined to process the same prompt during inference. 

To address this, we introduce the concept of column masks for each LoRA, denoted as $m_s$ and $m_c$. These column masks dynamically control the contribution of each column in the learned LoRA weights (see Fig.~\ref{fig:method}). Essentially, they allow the model to selectively activate or suppress specific columns within each LoRA, promoting orthogonality and reducing interference. 
By incorporating these column masks, the attention block update is modified as follows:
\begin{equation}
    K~\textnormal{or}~V(x, x_s, x_c) = W_0^T x +  m_s\Delta W_s^T x_s +  m_c\Delta W_c^T x_c.
\end{equation}

\begin{figure*}[t]
    \centering
    \includegraphics[width=1\linewidth]{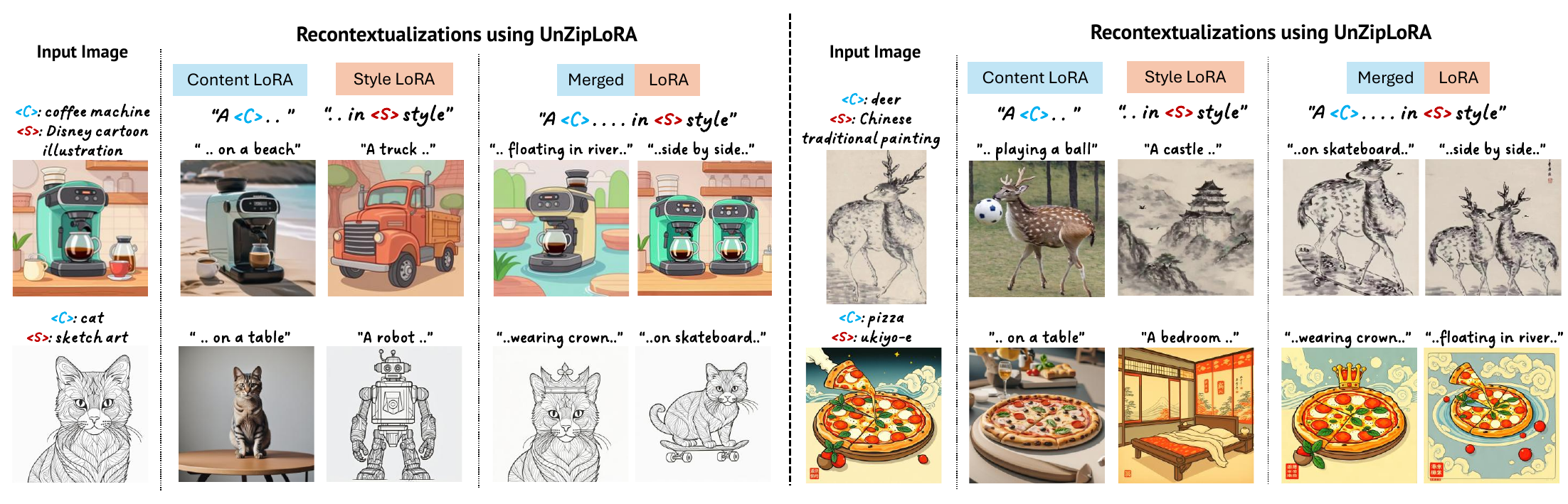}
    \vspace{-1.9em}
\caption{\textbf{Recontextualization. } The trained style and content LoRAs can be used individually or jointly at the inference time. The learned concepts can be used to generate images in various contexts, validating our method's robustness. Additional examples in the supplementary.}
    \vspace{-0.5em}
    \label{fig:recontext}
\end{figure*}

\begin{figure*}[t]
    \centering
    \includegraphics[width=\linewidth]{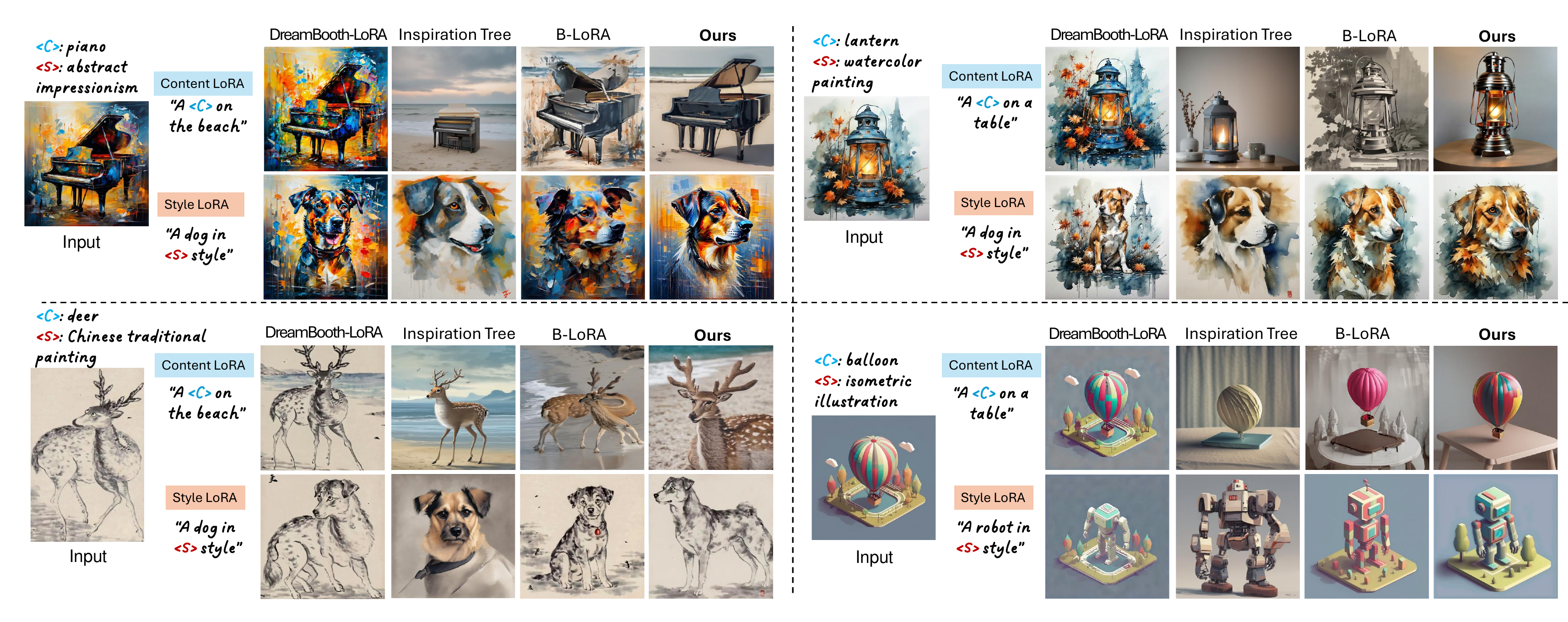}
    \vspace{-1.9em}
    \caption{\textbf{Qualitative Comparisons. }Subject and style decomposition outputs from UnZipLoRA compared against DreamBooth-LoRA~\cite{ruiz2023dreambooth}, Inspiration Tree~\cite{Vinker2023ConceptDF}, and B-LoRA~\cite{Frenkel2024ImplicitSS}. UnZipLoRA achieves superior subject and style fidelity, disentangling and preserving these concepts more effectively. More examples can be found in the supplementary.}
    \label{fig:sota_comparison}
    \vspace{-0.5em}
\end{figure*}
\noindent\textbf{Sparse masks with importance re-calibration.}
To further decompose the concepts, instead of training the entire LoRA matrices, we find that training only a fraction of total columns in each weight matrix is sufficient to learn the concepts, and it ensures weight sparsity for improved decomposition. This strategy is inspired by Liu et al.~\cite{liu2023cones}, who find that a small set of neurons tend to be much more salient than the others for capturing concepts. 

We use a dynamic approach to select the most important $N\%$ of the columns during training. Before the training, we initialize the column masks, $m_s$ and $m_c$, with the top $(N/3)\%$ of the most important columns for style and content. Importance is calculated using the Cone method~\cite{liu2023cones}, with five warm-up training steps using the full LoRA weights. Then, during the remainder of training, for every $t$ steps, we re-calibrate the column masks by calculating the column importance of LoRA weights and adding the new top $(N/3)\%$ of the most important columns to $m_s$ and $m_c$ until the $N\%$ cap is reached. In practice, $N$ in the range of $25$ to $40$ work well, and we choose $N=30$ for our experiments. 

\noindent \textbf{Orthogonal loss.} To promote compatibility between the subject and style LoRAs, we leverage 
the following orthogonality loss on $m_c$ and $m_s$:
\begin{equation}
\vspace{-0.2em}
    \mathcal{L}_{\perp} = \sum_{i} |m_c^i \cdot m_s^i|,
\vspace{-0.2em}
\end{equation}
minimizing which promotes orthogonality between the learned content and style weights~\cite{shah2024ziplora}. Orthogonal loss $\mathcal{L}_{\perp}$ is added to our reconstruction loss $\mathcal{L}_{DB}$ as regularizer with the weight parameter $\lambda_{\perp}$. 

This strategy brings significant improvements in compatibility of the resulting LoRAs, preventing overfitting to the input image and producing better recontextualizations, as shown in Fig.~\ref{fig:ablation} (row M2).

\subsubsection{Block Separation}\label{sec:block-separation}
B-LoRA~\cite{Frenkel2024ImplicitSS} showed that certain blocks in the SDXL U-Net are more responsible for content and some are more responsible for style. We can leverage this insight by relaxing the column sparsity constraints on the style and subject LoRAs corresponding to the style-sensitive and subject-sensitive blocks, respectively. In other words, all the LoRA columns in these blocks are fully trained without sparse masks. In our attempt to take advantage of B-LoRA’s block constraints, we find their split between subject and style is coarse-grained: tellingly, they fail to extract and represent the subject from the input image in its unstylized appearance. Therefore, we extend their approach by involving more blocks, and performing more fine-grained block-wise allocations of subject and style within the Up-blocks of SDXL U-Net (details in supplementary). This further improves the accuracy of fine details, especially for the style LoRA (see last row of Fig.~\ref{fig:ablation}).

\section{Experiments}
\label{sec:expt}
\subsection{Implementation Details}
\noindent \textbf{Dataset}. In our experiments, we use a set of $40$ diverse images with unique styles and subjects. These are collected from previous work, state-of-the-art text-to-image generators, and open-source repositories. Attribution information is provided in the supplementary material. 

\noindent \textbf{Experimental setup}. We use SDXL v1.0~\cite{Podell2023SDXLIL} as our base model. Subject and style LoRAs are trained with rank $=64$ using Adam (learning rate = $5e-5$) for $600$ steps with batch size $=1$, keeping the base SDXL weights and text encoders frozen. Column separation uses $t=200$, $N=30\%$, and weight of orthogonal loss is set to $\lambda_{\perp} = 0.5$ in all our experiments. Our block separation uses \emph{all} upsampling blocks in the SDXL U-Net, unlike B-LoRA that uses just two. See supplementary material for details of which blocks we assign to content and which ones to style learning.

\subsection{Qualitative Results}
As shown in Fig.~\ref{fig:teaser}, the subject and style LoRAs obtained by UnZipLoRA can be used separately to generate new representations of subject-only and style-only concepts. A key advantage of UnZipLoRA is its ability to produce compatible subject and style LoRAs that can be seamlessly merged via direct addition. This allows for the generation of novel recontextualizations that faithfully incorporate \emph{both} the subject and style of the original image. Fig.~\ref{fig:recontext} demonstrates this capability through recontextualizations using either individual LoRAs or a combination of both. More results and comparisons are included in the supplementary material.

\subsection{Comparative Evaluation}
\label{sec:comparitive evaluation}


In this section, we compare UnZipLoRA with three recent methods: DreamBooth-LoRA~\cite{ruiz2023dreambooth}, Inspiration Tree~\cite{Vinker2023ConceptDF}, and B-LoRA~\cite{Frenkel2024ImplicitSS}. As shown in Fig.~\ref{fig:sota_comparison}, our results are clearly superior qualitatively, and as a consequence, they are strongly preferred in a user study (Tab.~\ref{tab:userstudy}).

For DreamBooth-LoRA~\cite{ruiz2023dreambooth}, we train subject and style LoRA models separately using the DreamBooth method with the trigger phrases ``A \textless$c$\textgreater" and ``A \CP in \SP style." Note that the \CP and \SP used here are identical to those in UnZipLoRA. As can be seen from Fig.~\ref{fig:sota_comparison}, although DreamBooth-LoRA demonstrates satisfactory performance in learning the style, it fails to effectively destylize the subject and results in overfitting. This highlights the inherent difficulty in disentangling subject and style from a single image, even when training the LoRAs independently.

Inspiration Tree~\cite{Vinker2023ConceptDF} is a concept extraction method that decomposes the input into a binary concept tree using Textual Inversion~\cite{gal2022textual}. We train their model to extract subject and style using the prompt ``A \CP in \SP style", and follow their default setup that initializes the placeholders \CP and \SP with ``object" and ``art" respectively. Despite employing separate training for content and style, Inspiration Tree struggles to disentangle them accurately. While it can identify the overarching category of concepts correctly, it fails to capture the intricate details of the subject or the style. This limitation stems from its reliance on Textual Inversion, which focuses on learning text embeddings rather than fine-tuning the weights, resulting in limited expressiveness and controllability.

We also compare our approach with B-LoRA~\cite{Frenkel2024ImplicitSS} -- a technique aimed at combining style and subject of two different images. We follow the default setup of B-LoRA, and independently train both the subject and style representations on the same input image using same trigger phrases as UnZipLoRA. We keep other hyperparameters (learning rate, training steps, etc.) consistent with ~\cite{Frenkel2024ImplicitSS}. 
As evident in the lantern and balloon examples in Fig.~\ref{fig:sota_comparison}, B-LoRA outputs often retain residual style features and extraneous background elements, indicating overfitting and incomplete disentanglement. Furthermore, B-LoRA fails to accurately capture the color of the balloon, treating it as part of the style rather than the content. 
The limitations of B-LoRA can be attributed to its block-wise training approach, where content information is confined to specific blocks of the U-Net, potentially leading to information loss. By contrast, UnZipLoRA demonstrates more stable content destylization and better consistency in generating variations. 
Moreover, unlike B-LoRA, our method trains both the LoRAs jointly, reducing the compute requirements significantly. 

 Tab.~\ref{tab:userstudy} presents results of user studies comparing our method with the competing approaches for both subject/style decomposition and recontextualization. In our study, each participant is shown the input image, along with the outputs of two methods being compared (the methods are not labeled and their order is arbitrary). Each output group consists of $4$ images for the subject and $4$ images for style selected randomly, and participants are asked to choose an output group that decomposes the input image into style and content more accurately (see supplementary material for an example screenshot of the interface). We conducted three separate user studies -- one for each competing method vs. UnZipLoRA -- and received $204$ responses for each from $34$ total participants. As can be seen in Tab.~\ref{tab:userstudy},
UnZipLoRA is strongly preferred over all three competing methods for both decomposition and recontextualization. 

\begin{table}[t]
\begin{center}
\centering
\setlength{\tabcolsep}{4pt}
\captionof{table}{\small \textbf{User Preference Study}. We compare the user preference for subject-style decomposition (top), and combined subject-style recontextualization (bottom) between our approach and competing methods. Users generally prefer our approach in both. We received $204$ responses for each study from $34$ total participants. }
\label{tab:userstudy}
\vspace{-0.5em}
{\scalebox{0.74}{\begin{tabular}{cccc}
\toprule
\multicolumn{4}{c}{\textbf{$\%$ Preference for \name{} over:}} \\
\midrule
 & DreamBooth LoRA & \makecell{Inspiration Tree} & \makecell{B-LoRA}  \\
            \midrule
  \makecell{Decomposition} & $91.17\%$&$81.53\%$&$62.74\%$   \\
  \makecell{Recontextualization} &  $98.10\%$&$79.17\%$&$77.14\%$   \\
            \bottomrule
\end{tabular}}}


\captionof{table}{\small\textbf{Subject-alignment and Style-alignment Scores.} Comparisons for Content and Style Decomposition.}
\label{tab:alignscore}
{\scalebox{0.7}{\begin{tabular}{lcccc}
\toprule
 & \makecell{DB-LoRA} & \makecell{Insp. Tree} & \makecell{B-LoRA} & UnZipLoRA  \\
            \midrule
Style-align. (CLIP-I) $\uparrow$ & $0.417$ & $0.404$ &  $0.418$ & $\mathbf{0.427}$ \\
Subject-align. (DINO) $\uparrow$  & $0.339$ & $0.291$ &  $0.337$ & $\mathbf{0.349}$\\
\midrule
Style-align. (CSD) $\uparrow$  & $0.245$ & $0.229$ &  $0.244$ & $\mathbf{0.265}$ \\
Subject-align. (CSD) $\uparrow$  & $0.338$ & $0.334$ &  $0.342$ & $\mathbf{0.358}$\\ 
    \bottomrule
\end{tabular}}}
\vspace{-2.4em}
\end{center}
\end{table}

Tab.~\ref{tab:alignscore} further presents comparisons using automatic subject-alignment and style-alignment scores. We calculate alignment scores between decomposition output and the input image using standard CLIP-I image embeddings~\cite{radford2021clip} for style and DINO features~\cite{caron2021dino} for content. In both cases, we use cosine similarity as the metric and calculate averages over $8$ input images by generating $16$ samples for each. As can be seen from Tab.~\ref{tab:alignscore}, UnZipLoRA achieves the highest alignment scores for both subject and style, highlighting its superiority. DreamBooth-LoRA tends to overfit the input image, resulting in higher scores even though its outputs are qualitatively inferior to those of other methods.

CLIP-I and DINO metrics are inherently limited, especially in measuring style alignment, since they may not fully capture stylistic nuances and can be influenced by the semantic content of the images. As a more sensitive metric, we use a recently proposed CSD model~\cite{Somepalli2024MeasuringSS} trained specifically to extract the style descriptors from images. We use the embeddings from CSD's content and style branches to measure subject and style fidelity respectively. The CSD alignment scores (third and fourth line of Tab.~\ref{tab:alignscore}) yield a clearer separation between our method and the others.



\subsection{Ablation Study}
\label{sec:ablation}
\begin{figure}[t]
    \centering
    \includegraphics[width=\linewidth]{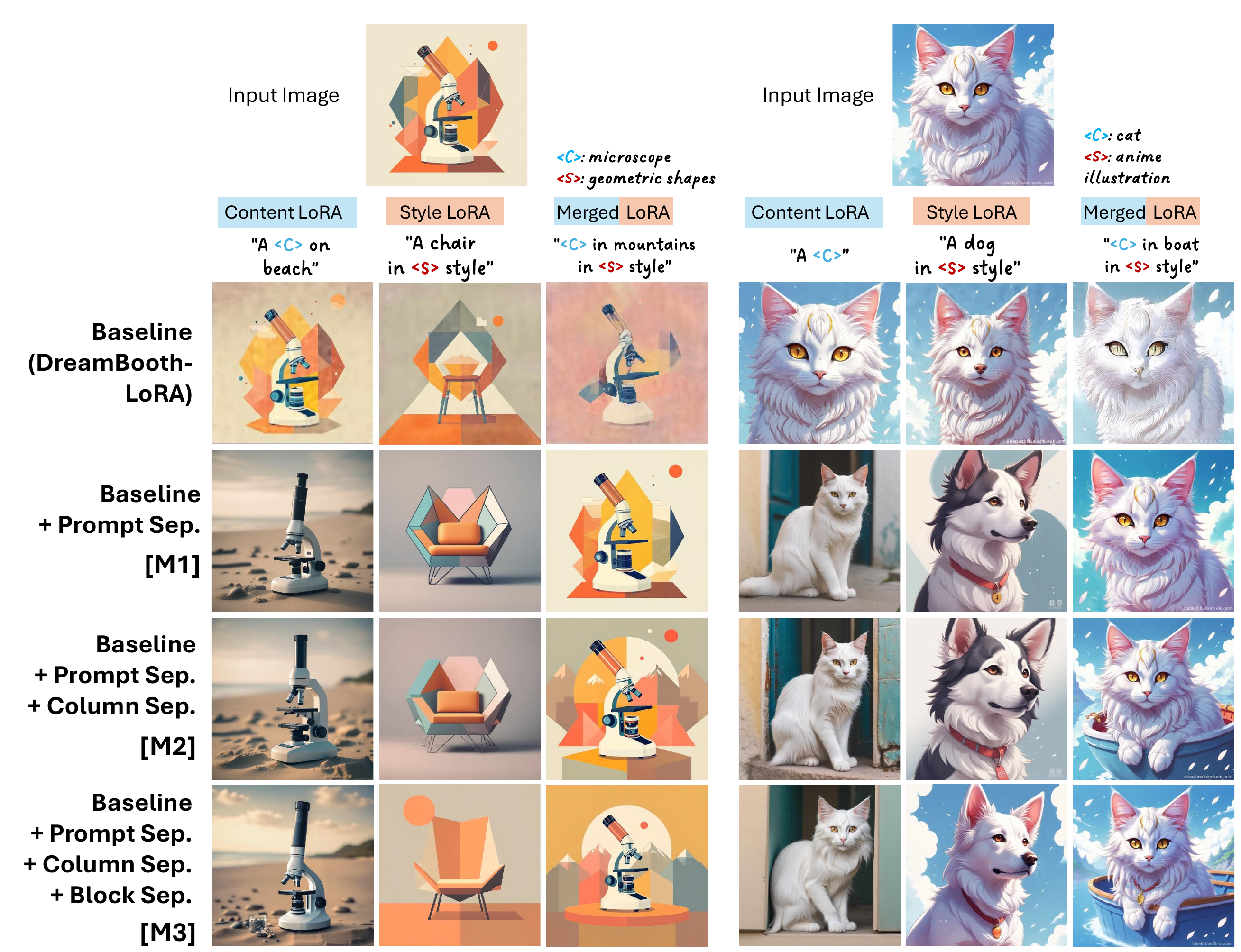}
    \caption{\textbf{Illustration of ablations.} Each column shows the performance of subject LoRA, style LoRA, and combined LoRA. 
    }
    \vspace{-0.8em}
    \label{fig:ablation}
\end{figure}

In this section, we present an ablation study to justify the effectiveness of our system design. As illustrated in Fig.~\ref{fig:ablation}, we start from DreamBooth~\cite{ruiz2023dreambooth} as the baseline and successively add the key components of our method: prompt separation (M1), column separation (M2), and block separation (M3). Tab.~\ref{tab:userstudy_ablation} reports user preferences for each model version over the previous one (based on $204$ responses from $34$ participants). 

As can be seen from Fig.~\ref{fig:ablation}, the DreamBooth baseline is incapable of subject-style disenganglement and has very weak recontextualization ability. 
Adding \textbf{prompt separation} (\textbf{M1}) to the baseline enables successful extraction of a realistic content LoRA and improves recontextualization. However, prompt separation alone struggles to capture all the complexities of the style from the input image (only 12.35\% user preference over baseline for style decomposition), and shows strong overfitting to the input image in combined recontextualization, indicating incompatibility of the learned LoRAs. 
Adding \textbf{column separation} (\textbf{M2}) on top of \textbf{M1} reduces the interference between subject and style LoRAs while also improving style generation abilities. For example, details like the color of the microscope in Fig.~\ref{fig:ablation} are retained, and the combined recontextualization performance improves significantly. 
However, improvements in style decomposition remain modest, indicating that using a small portion of the columns for style is insufficient for effective style learning. 
Finally, adding \textbf{block separation} (\textbf{M3}) with subject- and style-specific blocks on top of \textbf{M2} significantly enhances the model's ability to capture fine-grained stylistic features. The last row of Fig.~\ref{fig:ablation} shows that using all three separation strategies together achieves the best results, enabling comprehensive disentanglement of subject and style and flexible recontextualization. 




\begin{table}[t]
\begin{center}
\centering
\setlength{\tabcolsep}{4pt}
\captionof{table}{\textbf{Ablation User Study}. 
\textbf{Baseline}: Dreambooth-LoRA; \textbf{M1}: Prompt-separation only; \textbf{M2}: Prompt- and column- separations; \textbf{M3 (full)}: Prompt-, column-, and block- separations.}
\label{tab:userstudy_ablation}
\vspace{-0.9em}
\noindent
\centering
    {\scalebox{0.75}{\begin{tabular}{cccc}
    \toprule
&\multicolumn{3}{c}{\textbf{$\%$ Preference for:}} \\
\cmidrule(lr){2-4}
&\small{\makecell{\textbf{M1 over Baseline}}} & \small{\makecell{\textbf{M2 over M1}}} & \small{\makecell{\textbf{M3 over M2}}}  \\
                \midrule
      {\small\makecell{Subject Decomposition}} &$91.67\%$ & $55.74\%$ & $\mathbf{55.36\%}$ \\
      \midrule
       {\small\makecell{Style Decomposition}} &$12.35\%$ & $39.51\%$ & $\mathbf{86.42\%}$  \\
       \midrule
       {\small\makecell{Combined Recontextualization}}   &$92.80\%$ & $93.64\%$ & $\mathbf{61.90\%}$ \\
    \bottomrule
    \end{tabular}}}
\vspace{-2em}
\end{center}
\end{table}

\subsection{Applications and extensions} \label{sec:extensions}

\begin{figure}
    \centering
    \includegraphics[width=1\linewidth]{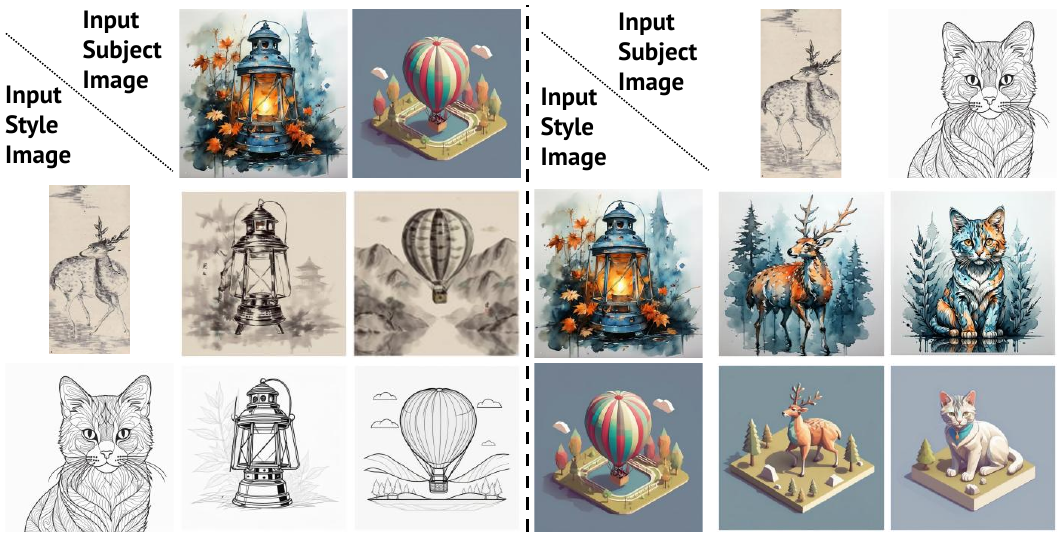}
    \vspace{-6.5mm}
    \caption{UnZipLoRA provides the valuable capability of \emph{cross-composition} using subject and style LoRAs from different images.}
    \label{fig:cross_comb}
    \vspace{-0.4em}
\end{figure}

\noindent\textbf{Cross-combination of subject and style LoRAs from different images. } The subject and style LoRAs produced by UnZipLoRA open up a possibility for cross-combination: pairing a subject LoRA from one image with a style LoRA from another. Fig.~\ref{fig:cross_comb} shows such cross-combination results where the LoRAs are combined by direct addition. While these LoRAs are not explicitly trained together (and thus not subject to the orthogonality constraints enforced by ZipLoRA~\cite{shah2024ziplora}), the inherent separation imposed by our column and block strategies generally results in higher compatibility than generic DreamBooth-LoRAs trained without such constraints. Consequently, direct arithmetic merger yields promising cross-stylization results.

\begin{figure}
    \centering
    \includegraphics[width=1\linewidth]{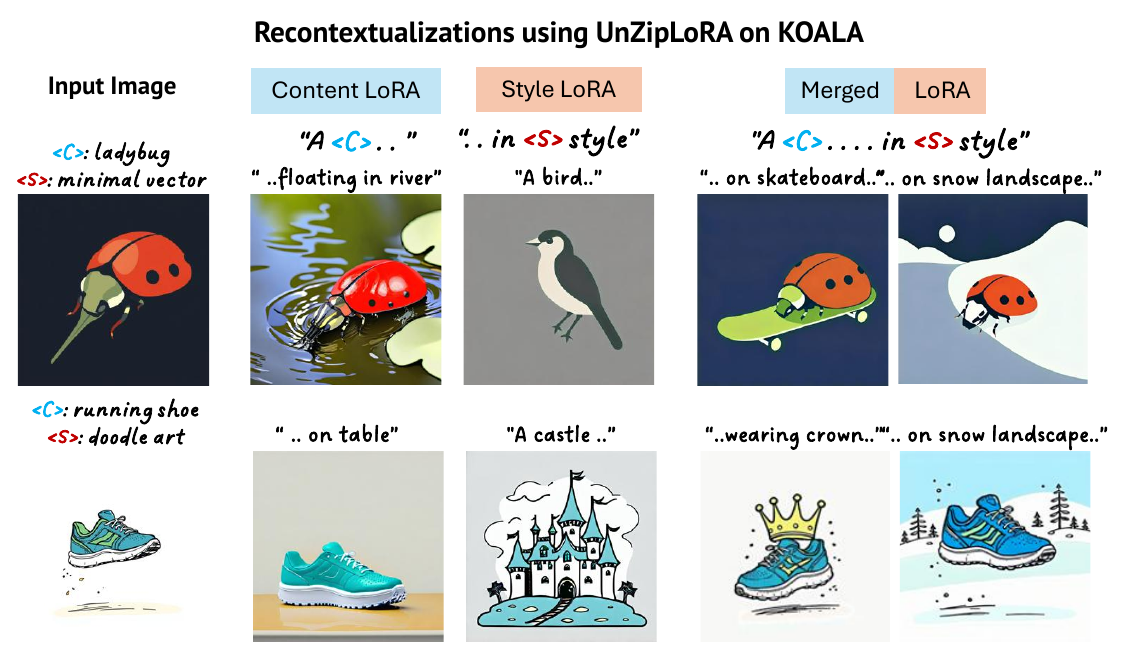}
    \vspace{-5mm}
    \caption{Our approach generalizes effectively, as demonstrated by successful results on the more recent KOALA diffusion model.}
    \label{fig:koala}
    \vspace{-1em}
\end{figure}

\noindent\textbf{Extension to other architectures. }To demonstrate the possibility of extending our method beyond SDXL, we train on KOALA~\cite{Lee@koala}, a more efficient recent model with a leaner U-Net architecture. As shown in Fig.~\ref{fig:koala}, our method, when applied to KOALA, accurately captures subject and style and allows for successful recontextualization (though the overall quality of the results is not as high as for SDXL due to limited capacity and lower parameter count of KOALA). The core idea of UnZipLoRA, that of simultaneously training two LoRAs on the same input image, should also be applicable to other architectures like DiT~\cite{Peebles2023DiT}, though this extension is beyond the scope of present work.

\begin{figure}[t]
    \centering
    \includegraphics[width=0.95\linewidth]{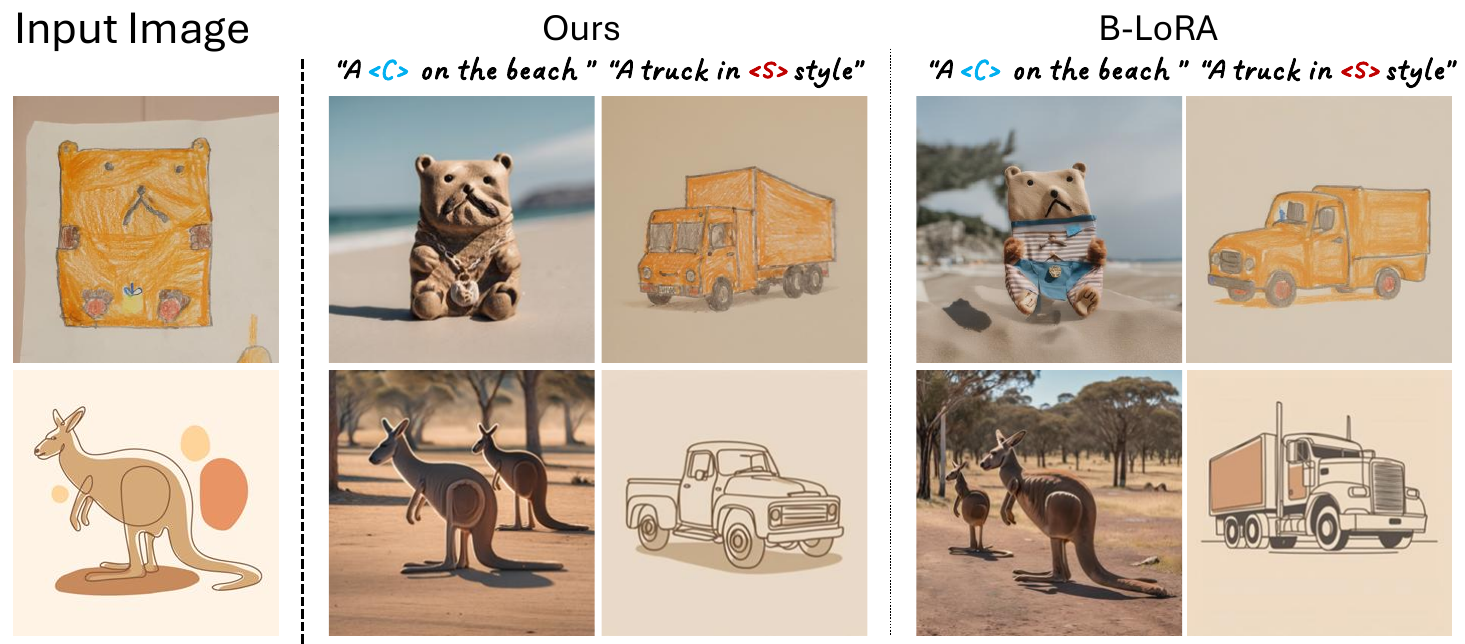}
    \caption{\textbf{Failure Cases. }In a few cases with highly abstract styles, UnZipLoRA may fail to destylize the subject accurately (note the unnatural shape of the bear). B-LoRA exhibits a similar failure.}
    \vspace{-1.2em}
    \label{fig:failure case}
\end{figure}

\section{Conclusion}
\label{sec:conclusion}
This paper introduced UnZipLoRA, a novel method for decomposing a single image into disentangled, compatible subject and style LoRAs. UnZipLoRA utilizes prompt, column, and block separation strategies to effectively extract these elements, enabling diverse recontextualizations and manipulations. Our experiments demonstrate superior performance compared to existing methods, highlighting UnZipLoRA's potential for creative exploration and control within text-to-image generation. While robust across a wide range of subjects and styles, UnZipLoRA may fail to destylize the subject when the style involves too much shape distortion or abstraction, as observed in Fig.~\ref{fig:failure case}. Future work includes exploring alternative disentanglement techniques for such challenging cases, training-free approaches for improved efficiency, and extending UnZipLoRA to other architectures for better generalization.


\vskip 1ex
\noindent \textbf{Acknowledgments.} This research was supported in part by NSF grant CCF 2348624. 

{
    \small
    \bibliographystyle{ieeenat_fullname}
    \bibliography{main, vsbib, vsbib2}
}

\clearpage
\resumetoc 
\renewcommand\thesection{\Alph{section}}
\renewcommand\thesubsection{\thesection.\arabic{subsection}}
\setcounter{section}{0}
\maketitlesupplementary
\setcounter{tocdepth}{3}
\tableofcontents
\section{Additional Implementation Details}
\label{sec:impl}
\begin{figure}[htbp]
    \centering
    \includegraphics[width=0.9\linewidth]{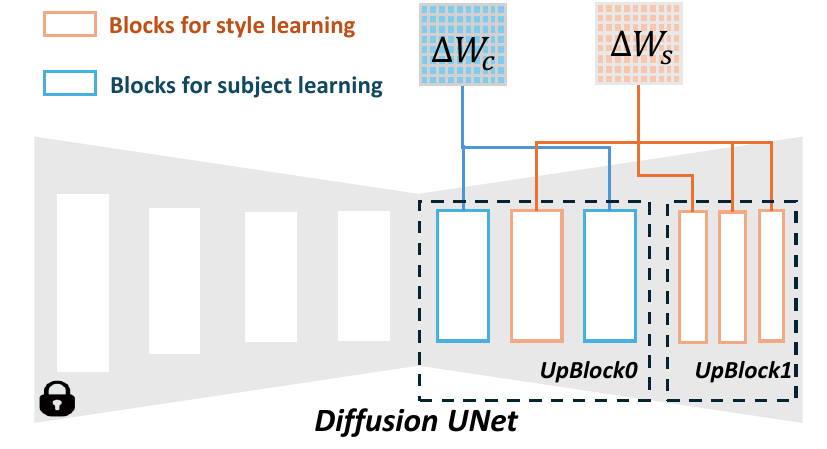}
    \caption{\textbf{Block separation strategy in the diffusion U-Net architecture of SDXL.} The illustration highlights the separation of transformer blocks into two distinct groups for disentangling content LoRA $\Delta W_c$ and style LoRA $\Delta W_s$.}
    \vspace{-0.8em}
    \label{fig:block_separation}
\end{figure}

In this section, we provide additional implementation details for our algorithm: 

\noindent \textbf{Block separation strategy.} As discussed in Sec.~\ref{sec:block-separation}, we employ a block separation technique similar to that proposed in B-LoRA. Specifically, as shown in Fig.~\ref{fig:block_separation}, the U-Net architecture in SDXL comprises three primary components: the downsampling blocks, the middle blocks, and the upsampling blocks. Each of these components contains several groups of transformer-based blocks. As the upsampling component is more critical to the overall performance, our primary focus for block separation lies in upsampling.  Within the upsampling component, there are two distinct groups of blocks, which are differentiated by number of transformers per block: the first group (\texttt{Upblock0} in Fig.~\ref{fig:block_separation}) contains blocks with 10 transformers each, while the second group (\texttt{Upblock1}) has blocks with only 2 transformers. Due to this disparity, the first group plays a more significant role in the learning process.

B-LoRA identifies that the first block of \texttt{Upblock0} mainly contributes for subject learning, while the second block of \texttt{Upblock1} is more specialized for style learning. This observation holds when the input images are relatively simple or lack intricate details. However, when the input images have fine-grained details, only one block is insufficient to capture all the necessary information. To enhance the content learning, we utilize additional block as well. Moreover, we leverage all the blocks in the second group of the U-net upblocks for style learning in order to maintain a balance of the learning capacities between the content and style. 
    
\noindent\textbf{Reproduction and experimental settings. }While reproducing the results of the competing methods, we use the exact hyperparameters reported in their respective papers (or their official implementations), including learning rate, training steps, and other experimental settings. For our approach, most experimental settings — such as the learning rate, batch size, and sampling frequency — are consistent with those described in the main paper. However, the number of training steps required for our method is $600$ for most input images, which is lesser as compared to other approaches.  Depending on the complexity of the input however, our method may require higher number training steps (in the range of $800$ steps) if input image contains fine-grained details.


\section{Additional Experiments and Results}
\label{sec:add_exp}
\subsection{Trigger Phrases Selection}
As mentioned in the main paper, we follow the standard prompt construction strategy ``a \CP in \SP style" with trigger phrases \CP and \SP in our text prompt to describe the content and style respectively. For selecting the trigger phrases \CP and \SP, we follow the descriptor strategies of existing approaches such as DreamBooth~\cite{ruiz2023dreambooth}, StyleDrop~\cite{sohn2023styledrop}, ZipLoRA~\cite{shah2024ziplora},  and B-LoRA~\cite{Frenkel2024ImplicitSS}.

\begin{figure*}[htbp]
    \centering
    \includegraphics[width=1.0\linewidth]{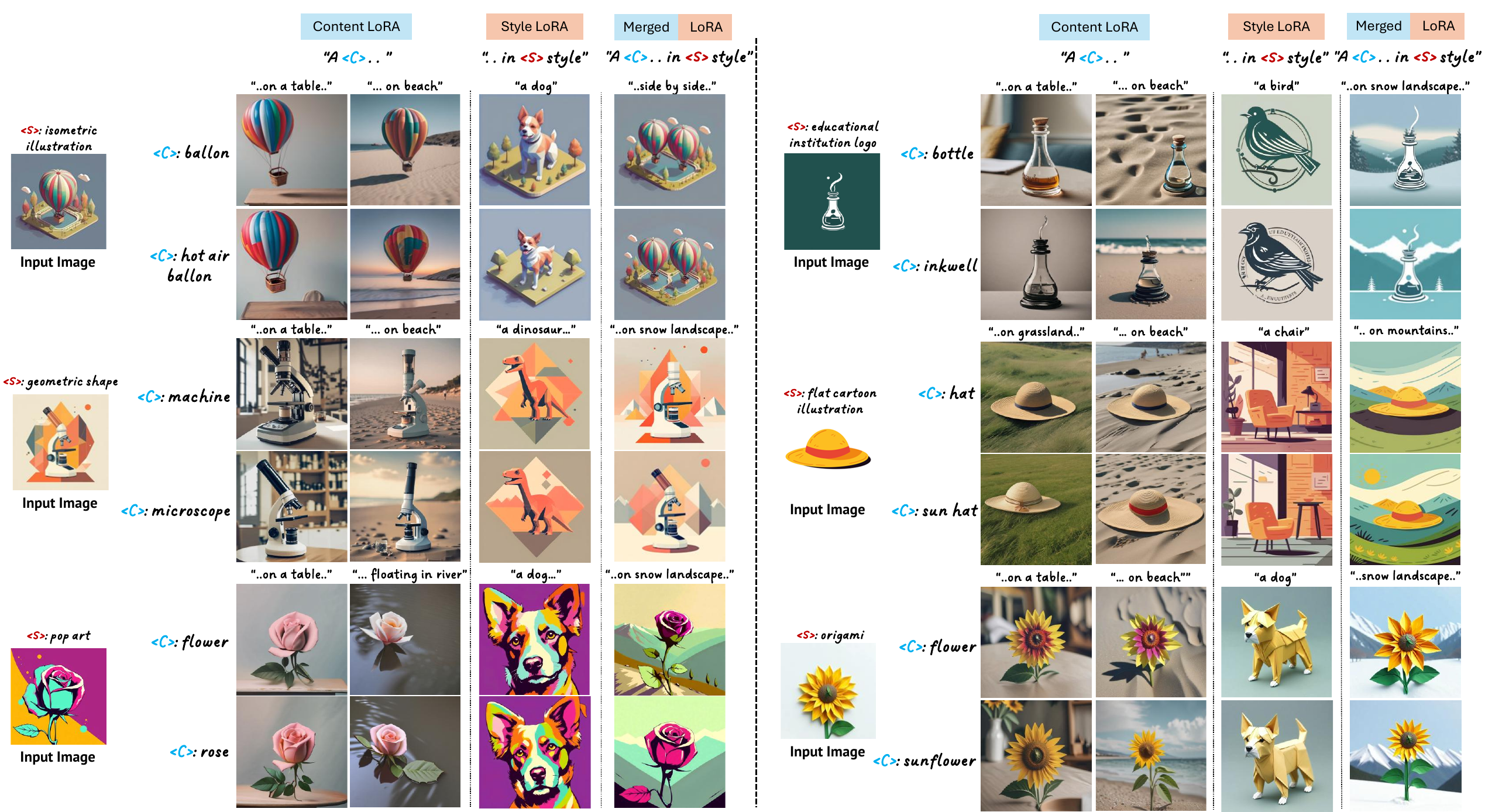}
    \caption{\textbf{Trigger phrase selection for the subject}. }
    \vspace{-0.5em}
    \label{fig:prompt_sel_subject}
\end{figure*}

\begin{figure*}[htbp]
    \centering
    \includegraphics[width=1.0\linewidth]{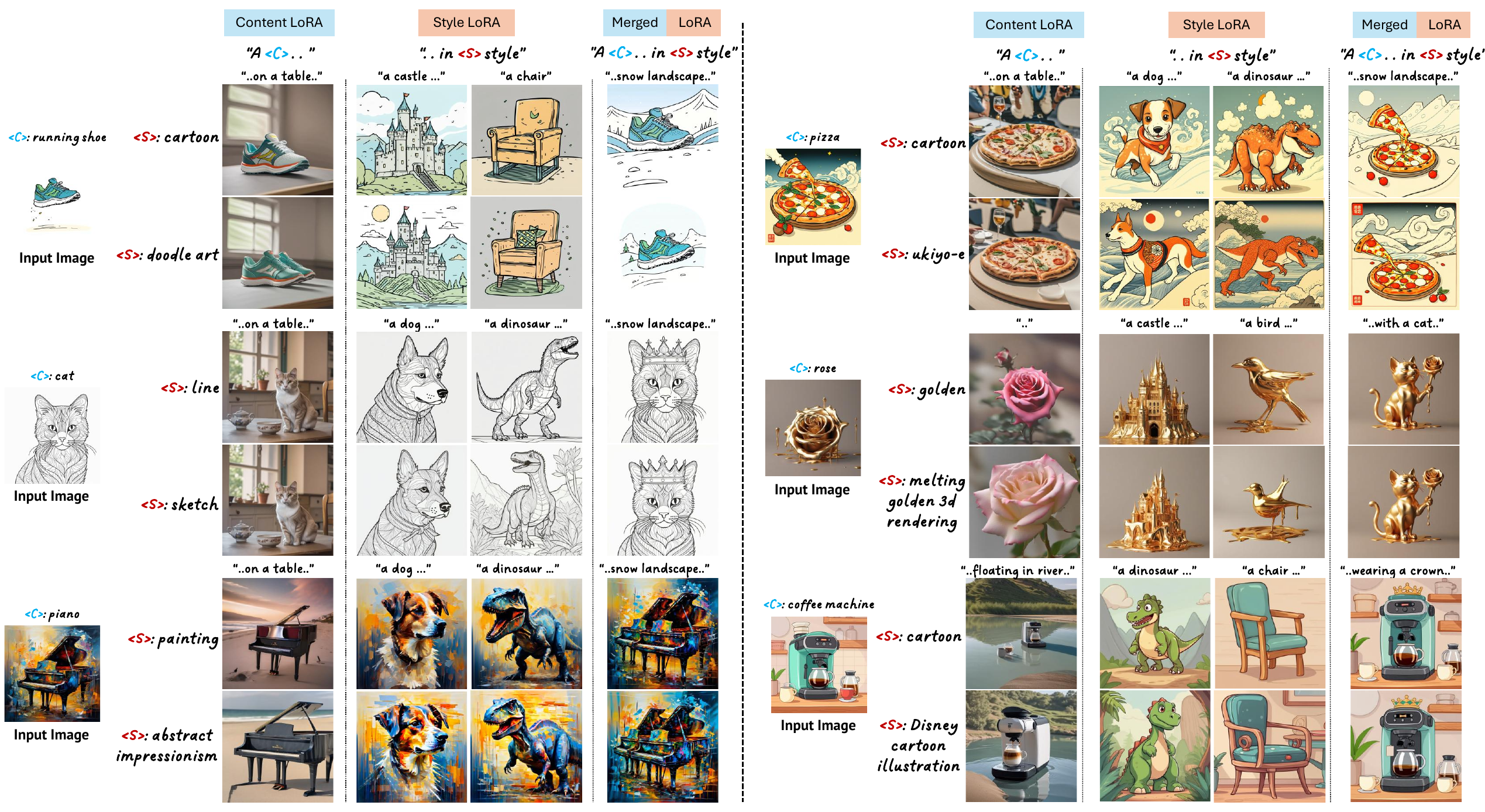}
    \caption{\textbf{Trigger phrase selection for the style}. }
    \vspace{-0.5em}
    \label{fig:prompt_sel_style}
\end{figure*}

\begin{figure*}[htb]
    \centering
    \includegraphics[width=\linewidth, trim=0 0 0 9mm, clip]{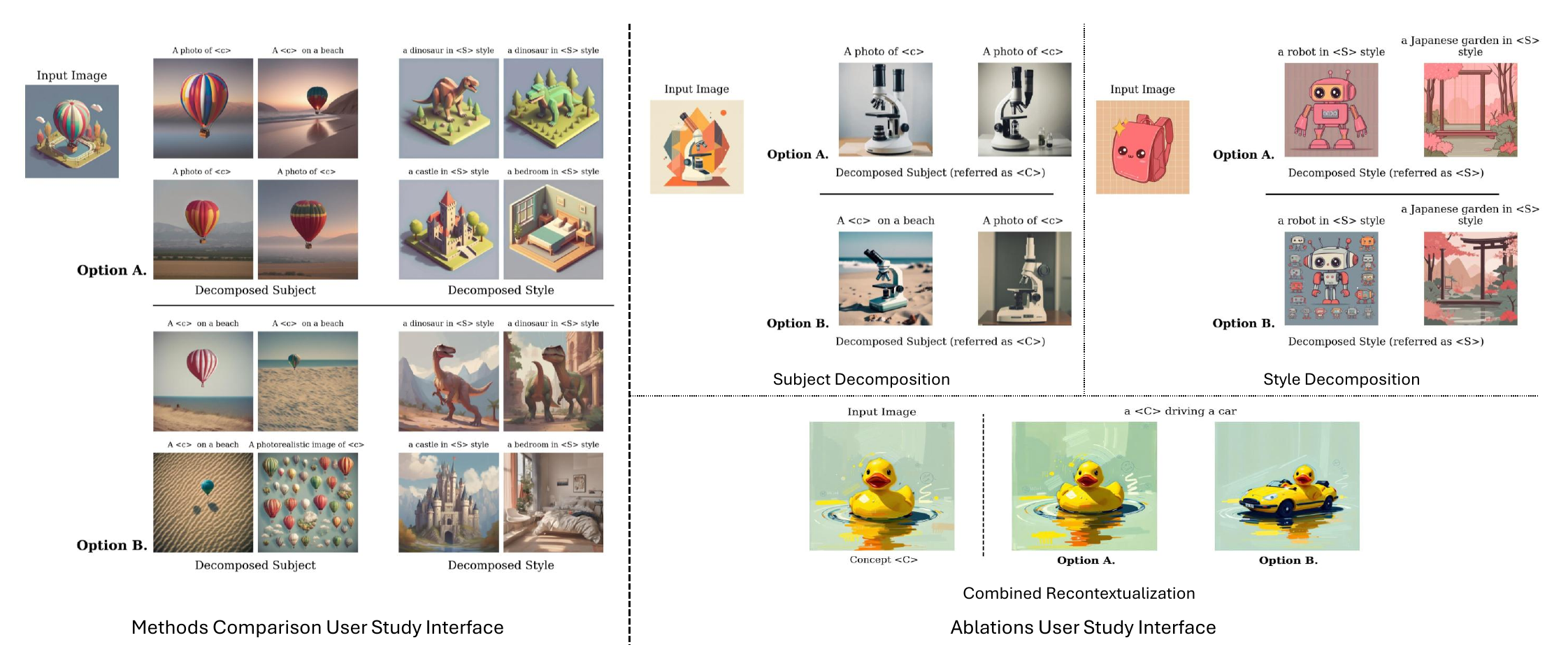}
    \caption{\textbf{User Study Interfaces. }We depict the graphical user interfaces (GUI) we used in (left) methods comparison user study for comparing quality of subject-style decomposition, and (right) ablations user study for comparing subject decomposition, style decomposition, and combined recontextualization. User study results are included in the main paper.}
    \label{fig:userstudy_interface}
\end{figure*}

\begin{figure*}[t]
    \centering
    \includegraphics[width=1.0\linewidth]{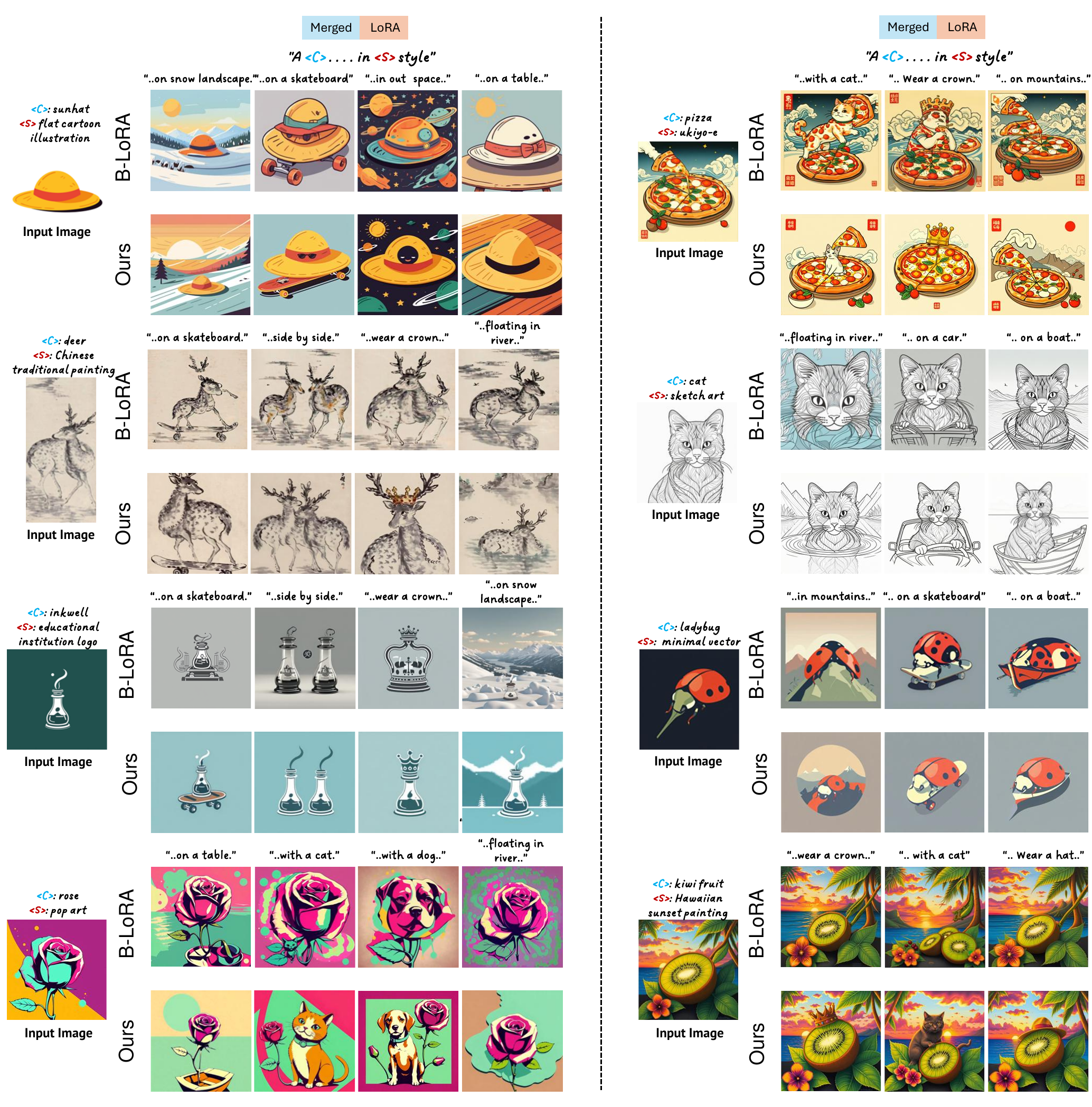}
    \caption{\textbf{Comparison of Subject-style Recontextualization}. We present a comparison of subject-style recontextualization between our method and B-LoRA across diverse prompts and input images. The results highlight our method's superior ability to flexibly adapt subjects and styles to various contexts while accurately reproducing both subject and style features.}
    \vspace{-0.8em}
    \label{fig:recontext_suppl}
\end{figure*}

\begin{figure*}[htbp]
    \centering
    \includegraphics[width=\linewidth]{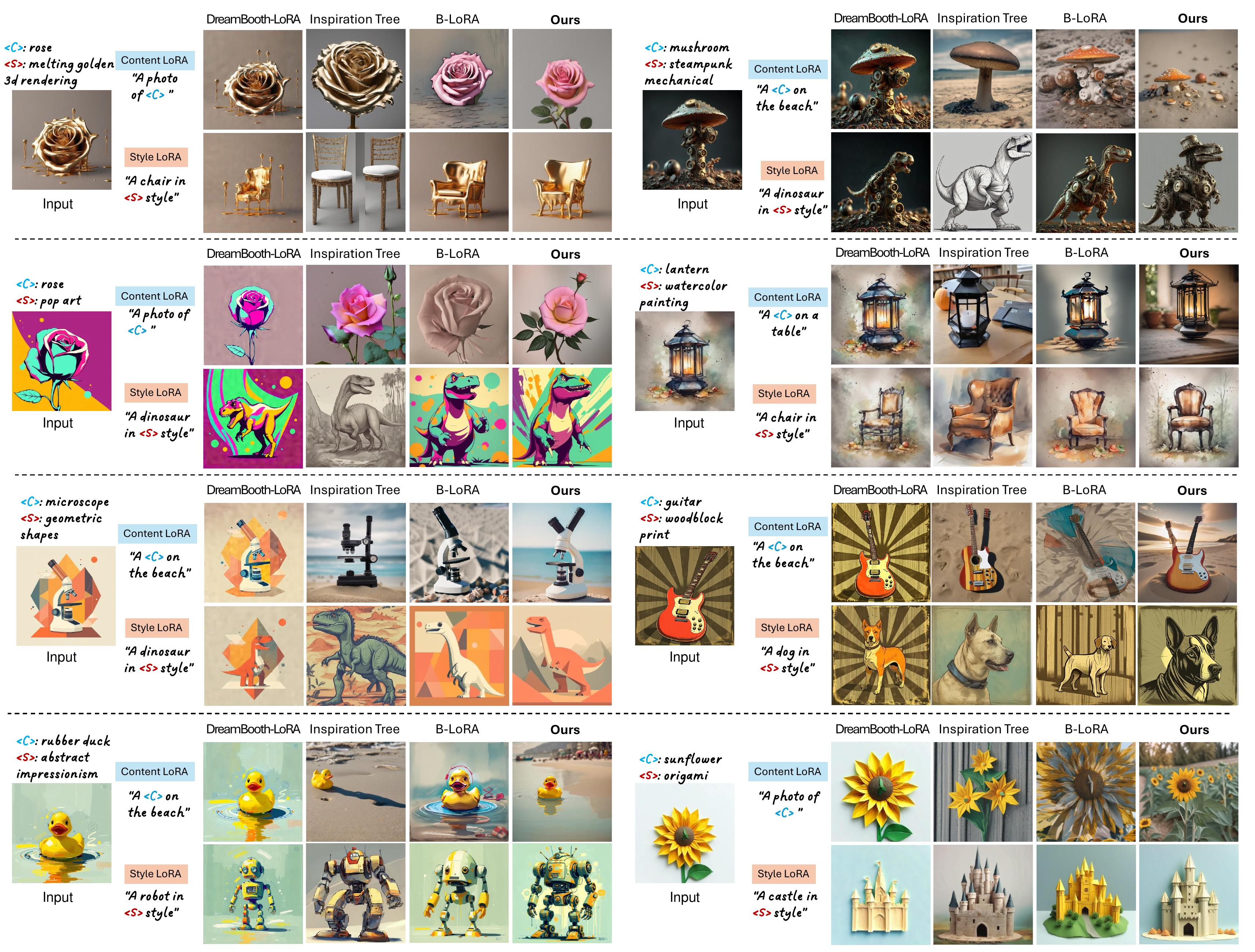}
    \caption{\textbf{Qualitative Comparisons. }Additional groups of compare subject and style disentanglement from ours method against DreamBooth-LoRA, Inspiration Tree, and B-LoRA. The result again demonstrates our superior ability to preserve the intended features compared to other methods.}
    \vspace{-0.8em}
    \label{fig:add_comparison}
\end{figure*}

As discussed in DreamBooth~\cite{ruiz2023dreambooth}, using a unique token for the subject helps the model to bind the new subject to a unique embedding vector in space since a subject is usually spatially localized in a particular region of the image. On the other hand, style is typically spread throughout the image, thus using just a generic text description increases the flexibility during style training as reported by StyleDrop~\cite{sohn2023styledrop}. This choices are further validated by the follow up works such as ZipLoRA~\cite{shah2024ziplora} and B-LoRA~\cite{Frenkel2024ImplicitSS}. Thus, in our experiments we use a unique token followed by the subject class label as the subject trigger phrase (e.g. \CP =`sks dog'), and use generic artistic description for the style (e.g. \SP = `watercolor painting'). For example, the prompt we use for the first input image in Fig.~\ref{fig:recontext_suppl} is ``A sks sun hat in flat cartoon illustration style". 

Here we independently validate the above choices by studying the impact of the trigger phrases on the results of our method as the degree of detail of these phrases is varied. We confirm that a single-word class label for the subject, and a generic, brief (2-3 word) description for style are sufficient to effectively guide our method, while a more detailed prompt provides additional flexibility for reinforcing any desired attributes.

In Fig.~\ref{fig:prompt_sel_subject} and Fig.~\ref{fig:prompt_sel_style}, we show several groups of examples demonstrating the influence of subject and style trigger phrases on generation. We compare results trained with more general prompts, such as referring to a subject by its category rather than its specific name or using broad and vague style phrases (e.g. using `flower' instead of `sunflower'). The results indicate that the generation quality with general prompts is largely preserved, showing no noticeable degradation compared to more detailed prompts. 

In some cases, using detailed phrases can help reinforce specific attributes, thus providing flexibility to users. For example, in Fig.~\ref{fig:prompt_sel_subject}, using `flower' as a trigger phrase retains most of the characteristics of the input subject, while using a more detailed prompt `sunflower' helps boost the fidelity of the features in the center of the flower. Similarly, using `microscope' instead of generic `machine' helps retain the fine-grained shape and color characteristics of its eyepiece tube.

Similar conclusions hold for style trigger phrases as well (see Fig.~\ref{fig:prompt_sel_style}): UnZipLoRA can successfully learn the style of the input image even with highly generic, single word style phrases such as `cartoon' and `painting'. At the same time, more detailed descriptions can help reinforce specific attributes. While these attributes may still be captured without explicit mention, incorporating them into the prompt ensures more stable preservation. For example, the differences in generations of `golden' style in Fig.~\ref{fig:prompt_sel_style} are subtle, yet a more artistically descriptive phrase `melting golden 3d rendering' leads to clearer stylistic features such as flowing golden droplets. Another example is the pizza in Fig.~\ref{fig:prompt_sel_style}: UnZipLoRA works well with a generic phrase `cartoon', and specifying `ukiyo-e' helps retaining more of the background waves, and produces a stronger `ukiyo-e' aesthetics.

These findings suggest that general trigger phrases suffice for capturing the overall subject and stylistic impression. When retaining specific features is required, explicitly incorporating those features into the input prompt is beneficial.

\subsection{Subject-Style Recontextualization Comparison}
\label{sec:suppl_recontext}

A key advantage of UnZipLoRA is its ability to produce compatible subject and style LoRAs that can be seamlessly merged via direct addition. This allows for the generation of novel
recontextualizations that faithfully incorporate both the subject and style of the original image. Figure~\ref{fig:recontext} in the main paper indemonstrates this
capability through various recontextualizations using either
individual LoRAs or a combination of both. In this section, we provide qualitative comparisons between our method and B-LoRA for the task of  subject-style recontextualization. As demonstrated in Fig.~\ref{fig:recontext_suppl}, UnZipLoRA is superior in preventing overfitting, reproducing accurate subject and style representations, and enabling flexible recontextualization.

\noindent \textbf{Preventing overfitting. }Our method mitigates overfitting by disentangling subject and style representations, ensuring diverse and robust outputs even with challenging prompts

\noindent \textbf{Accurate subject and style reproduction. }We achieve precise reproduction of the input's subject and style elements while avoiding blending artifacts.

\noindent \textbf{Flexible recontextualization. }Our method enables diverse and logical recontextualization, handling both straightforward prompts like "on a table" and complex, creative prompts that require a nuanced extraction of subject and style.

We conducted user studies in the main paper to compare our method with the competing approaches. Beyond the configurations, results, and analyses presented in the main paper, we include the interface used for the main user study and the ablation user study in Fig.~\ref{fig:userstudy_interface}.

\begin{figure*}[tb]
    \centering
    \includegraphics[width=1.0\linewidth]{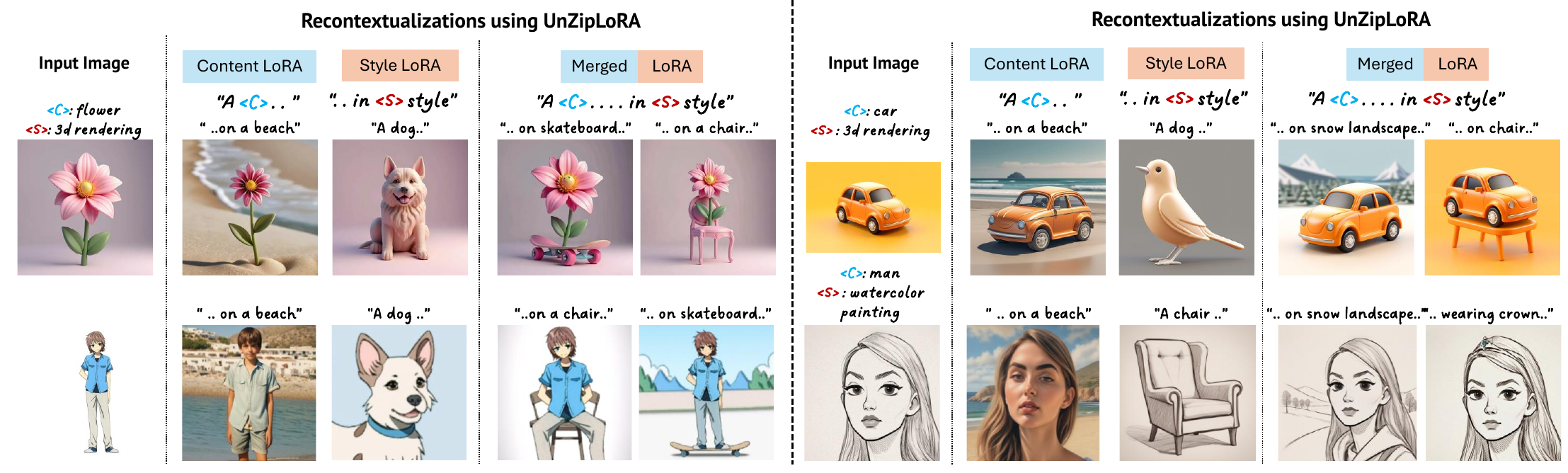}
    \caption{\textbf{Additional Decomposition and Re-contextualization Results. }We present additional results on diverse inputs such as humans and 3D rendering styles. The results showcase UnZipLoRA's ability to preserve the details of both the subject and style along with providing flexible recontextualizations.}
    \vspace{-0.8em}
    \label{fig:extra_suppl}
\end{figure*}

\subsection{Additional Qualitative Comparisons}

To complement the findings in the main paper, we provide additional qualitative comparisons in Fig.~\ref{fig:add_comparison} across more diverse prompts and input images. These examples further demonstrate the superior performance of our method in subject-style disentanglement, detail preservation, and overfitting prevention compared to DreamBooth-LoRA, Inspiration Tree, and B-LoRA.

Upon closer inspection of the examples, our observations are consistent with the results presented in the main paper. Each of the compared methods demonstrates limitations that prevent them from achieving the level of disentanglement and flexibility required for our task.

\noindent \textbf{DreamBooth-LoRA. } DreamBooth-LoRA struggles to disentangle subject features from style, even though it captures some stylistic features effectively. However, its results often suffer from overfitting to the input image, limiting its ability to recontextualize the style in diverse settings. Our methods successfully captures the style with high fidelity, enabling flexible style recontextualization without overfitting to the input.

\noindent \textbf{Inspiration Tree. }While Inspiration Tree effectively prevents the overlap of subject and style concepts and consistently destylizes the input, it struggles to distinguish detailed features of both subject and style. This limitation results in outputs that lack the intricate details of the input subject or style. By incorporating separation strategies, our method intelligently learns and distinguishes these features, leading to more detailed and accurate outputs.

\noindent \textbf{B-LoRA. } As what we discussed in Experiments section of main paper, and in Sec.~\ref{sec:suppl_recontext}, B-LoRA fails to generate consistent results and suffers from overfitting to input images. While it can always accurately learn style features, it struggles to reproduce subject details reliably. For instance, in the guitar and sunflower examples, B-LoRA fails to consistently retain the original input's color.
    
Moreover, it often mixes subject and style features, resulting in generations that incorporate unintended elements, such as the background color of the microscope or the sunflower’s color being treated as part of the style. In contrast, our method addresses these issues with carefully designed separation techniques, ensuring consistent, disentangled outputs that faithfully represent both subject and style.

\subsection{Additional Qualitative Results on Diverse Inputs}
We present additional decomposition and recontextualization results using a wider variety of subjects/styles, including more complex subjects such as humans and styles such as 3D rendering in Fig.~\ref{fig:extra_suppl}. The results demonstrate that our method generalizes effectively to detailed and diverse subjects, maintaining the fidelity of the subjects/styles, and flexibly recontextualizing them across various contexts, showcasing the superiority of our method in handling rich and challenging inputs.

\subsection{Additional Ablation Study Results}
\begin{figure*}[tb]
    \centering
    \includegraphics[width=0.95\linewidth]{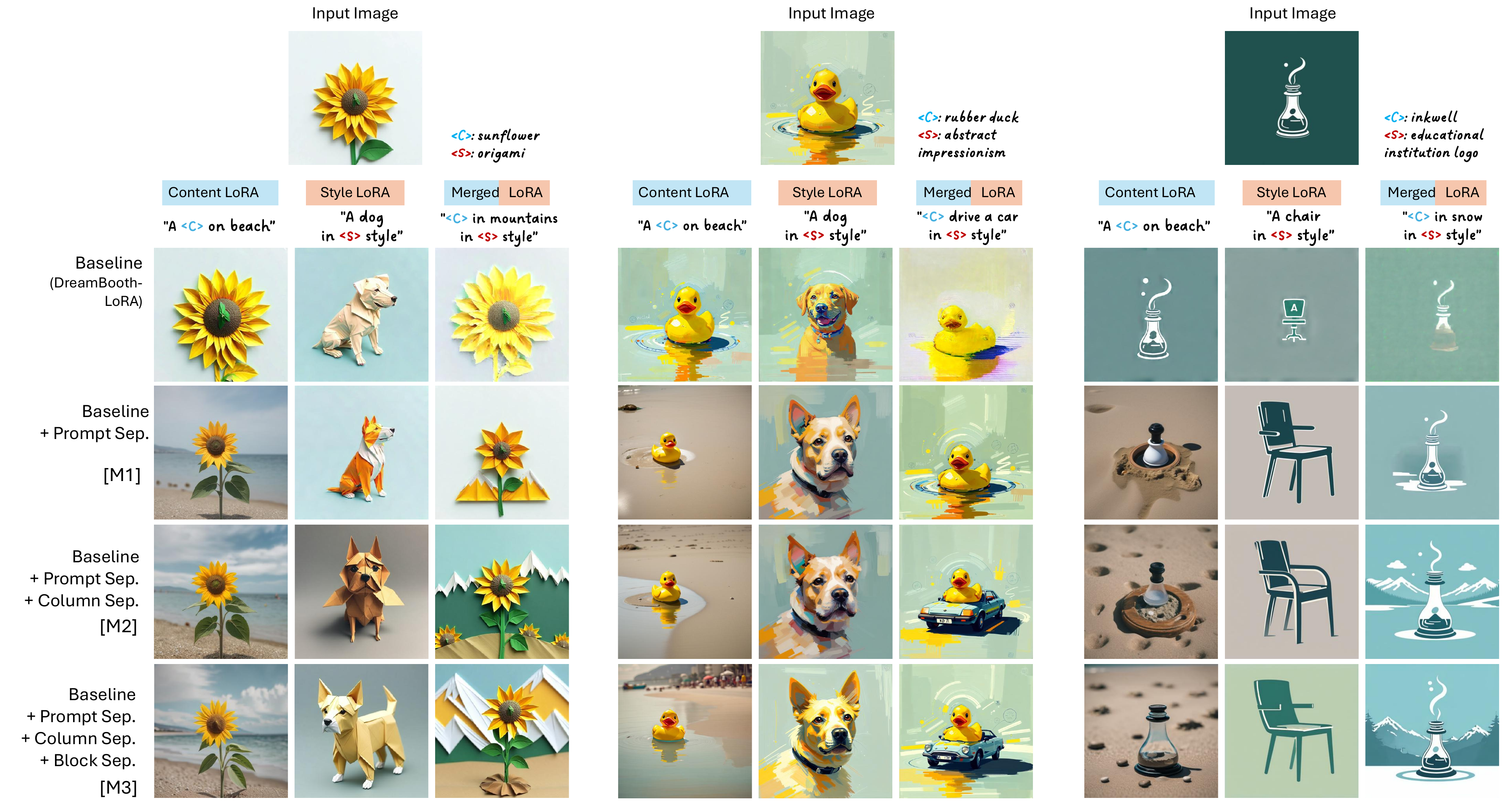}
    \caption{\textbf{Ablation study. }We show the impact of adding prompt-wise separation, column-wise separation, and block-wise separation sequentially. Each row illustrates the outputs of baseline methods and our proposed approaches, highlighting their difference in subject-style disentanglement, style fidelity, and recontextualization across different examples. }
    \vspace{-0.8em}
    \label{fig:extra_ablation}
\end{figure*}

\begin{figure*}[htbp]
    \centering
    \includegraphics[width=\linewidth]{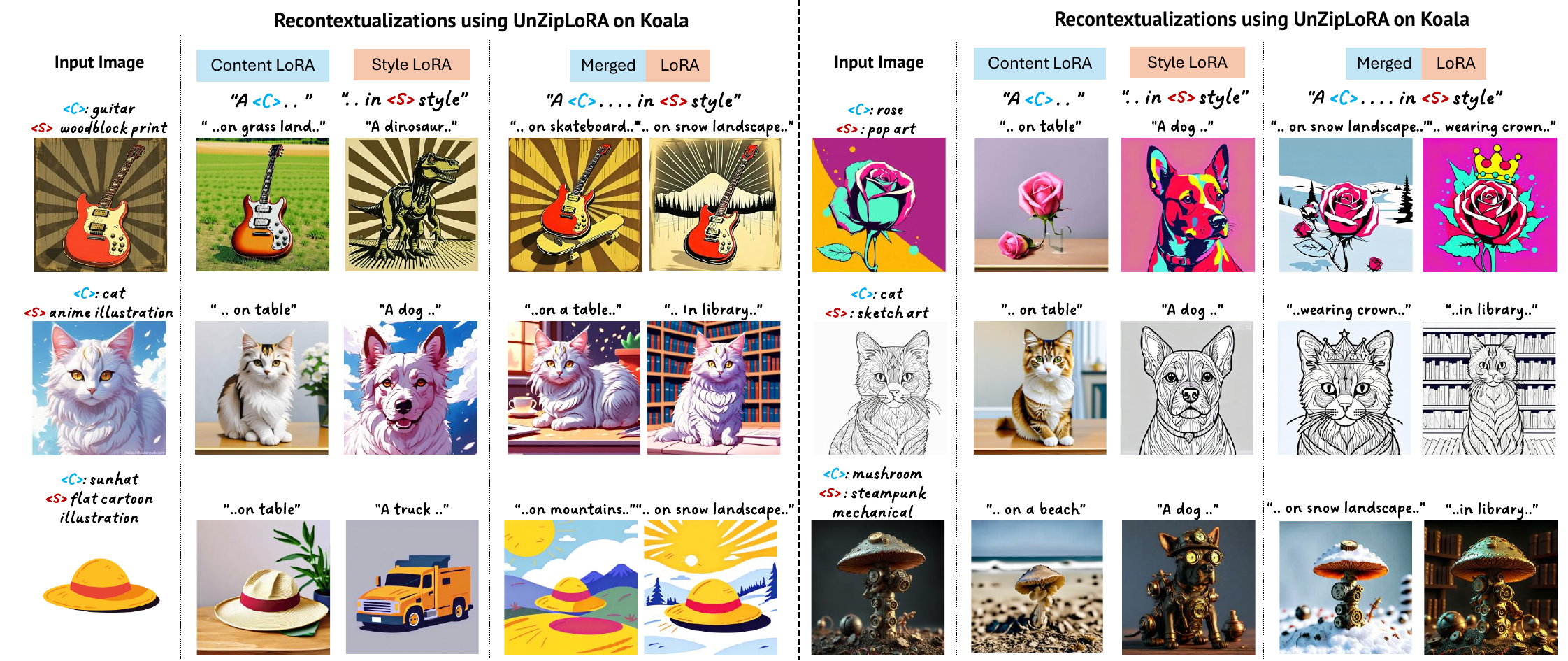}
    \caption{\textbf{Additional Results on KOALA Diffusion.} Our approach generalizes effectively, as demonstrated by successful results on the more recent KOALA diffusion model.}
    \vspace{-0.8em}
    \label{fig:add_kuala_suppl}
\end{figure*}

We provide additional examples of ablation study of the effects of adding each component, as shown in Fig.~\ref{fig:extra_ablation}. Each row demonstrates the recontextualization of subject, style and subject-style across different configurations and baseline methods. 

\noindent \textbf{Prompt separation. }Distinct prompts and LoRA weights for subject and styles guide their respective LoRAs, ensuring effective and disentangled subject-style decomposition.

\noindent \textbf{Column separation. }Dynamic column masks selectively activate relevant columns during training, preserving learning capacity with fewer columns and preventing interference. Enhanced orthogonality between LoRAs improves flexibility in recombination.

\noindent \textbf{Block separation. }Style-sensitive blocks effectively capture essential stylistic features, while subject-sensitive blocks focus on fine details. 
\begin{table}[htb]
\begin{center}
\centering
\caption{{\small\textbf{Ablation Study Alignment Scores.} Comparisons for Content and Style Decomposition among each separation strategy.}}
\label{tab:quantitive_ablation}
{\scalebox{0.72}{\begin{tabular}{lcccc}
\toprule
 & \makecell{DB-LoRA} & \makecell{\textbf{M1}} & \makecell{\textbf{M2}} & \textbf{M3 (UnZipLoRA)} \\
            \midrule
Style-align. (CLIP-I) $\uparrow$ & $0.417$ & $0.409$ &  $0.407$ & $\mathbf{0.427}$ \\
Subject-align. (DINO) $\uparrow$  & $0.339$ & $0.346$ &  $0.347$ & $\mathbf{0.349}$\\
\midrule
Style-align. (CSD) $\uparrow$  & $0.245$ & $0.217$ &  $0.216$ & $\mathbf{0.265}$ \\
Subject-align. (CSD) $\uparrow$  & $0.338$ & $0.352$ &  $0.354$ & $\mathbf{0.358}$\\ 
    \bottomrule
\end{tabular}}}
\end{center}
\end{table}

\noindent \textbf{Quantitative results. }Beyond the ablation user study results provided in Tab.~\ref{tab:userstudy_ablation} in the main paper, we provide additional quantitative results in the form of subject- and style-alignment scores in Tab.~\ref{tab:quantitive_ablation}. 
The results confirm each separation strategy’s independent contribution: prompt separation (M1) aids subject learning, column separation (M2) improves disentanglement, and block separation (M3) significantly enhances stylization. These trends align with the user study results (Tab.~\ref{tab:userstudy_ablation}) and our analysis in Sec.~\ref{sec:ablation} of the main paper.

\noindent \textbf{User study interface.} We conducted user studies in the main paper to compare our method with the competing approaches. Beyond the configurations, results, and analyses presented in the main paper, we include the interface used for the main user study and the ablation user study in Fig.~\ref{fig:userstudy_interface}.

\begin{figure}[htbp]
    \centering
    \includegraphics[width=\linewidth]{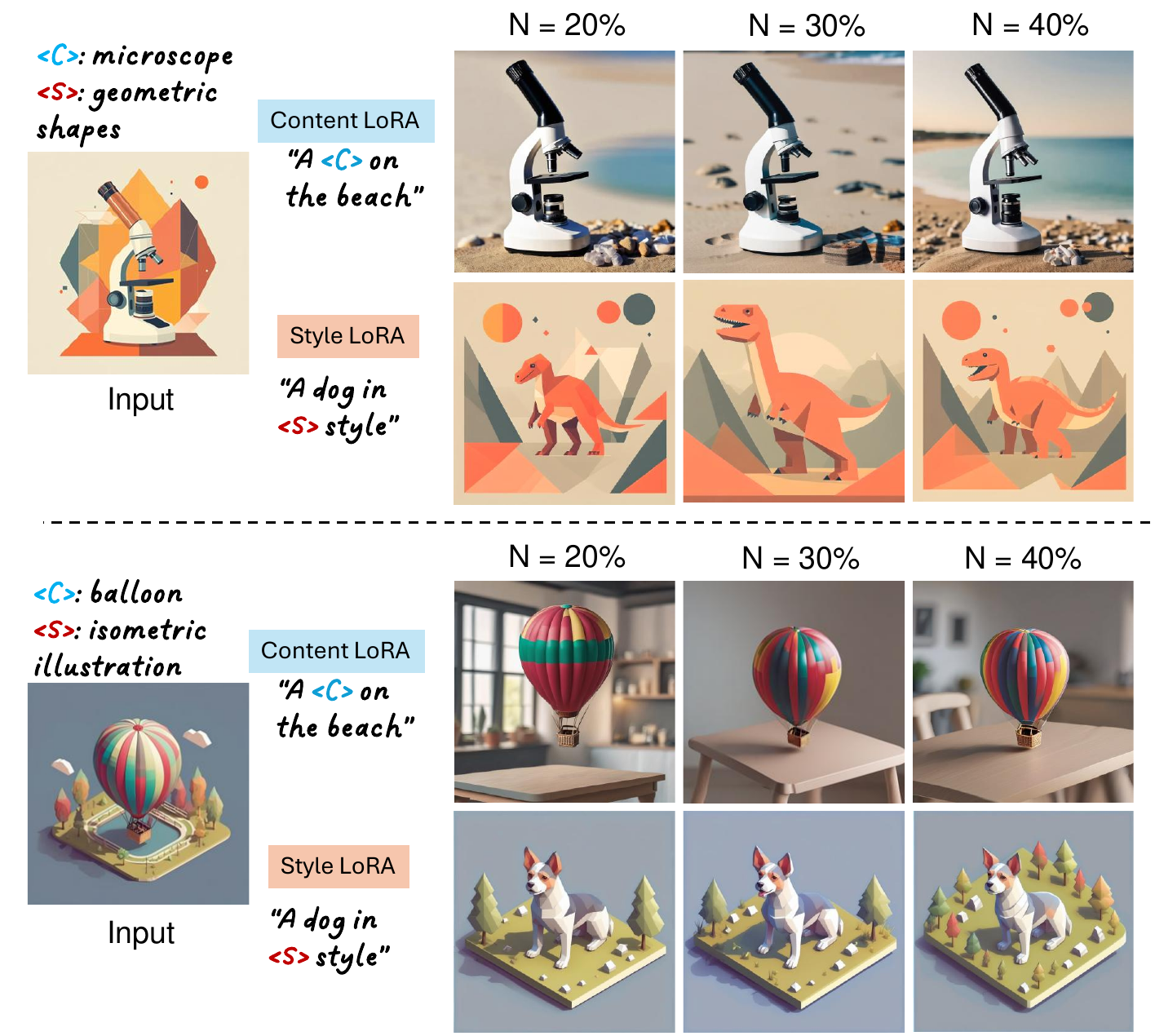}
    \caption{Ablations on different values of $N$. Our column selection strategy selects top $N\%$ of most important columns usiing dynamic importance recalibration strategy. As shown, $20\leq N\leq 40$ is sufficient for capturing subject/style successfully.}
    \vspace{-0.8em}
    \label{fig:add_column_sel}
\end{figure}

\subsection{Ablation on Percentage of High-importance Columns}
To evaluate the influence of the column selection percentage (N) within our dynamic importance recalibration strategy, we conducted ablation studies across varying values of $N$. This strategy selects the top $N\%$ of most important columns by calculating the column importance from the gradient information using the Cone method~\cite{liu2023cones}. The results of our ablation experiments shown in Fig.~\ref{fig:add_column_sel} confirm that a selection percentage within the range of $20\leq N\leq 40$ is sufficient for successfully capturing both subject and style characteristics.

\subsection{Additional Cross-combination Results}
\begin{figure}[htbp]
    \centering
    \includegraphics[width=\linewidth]{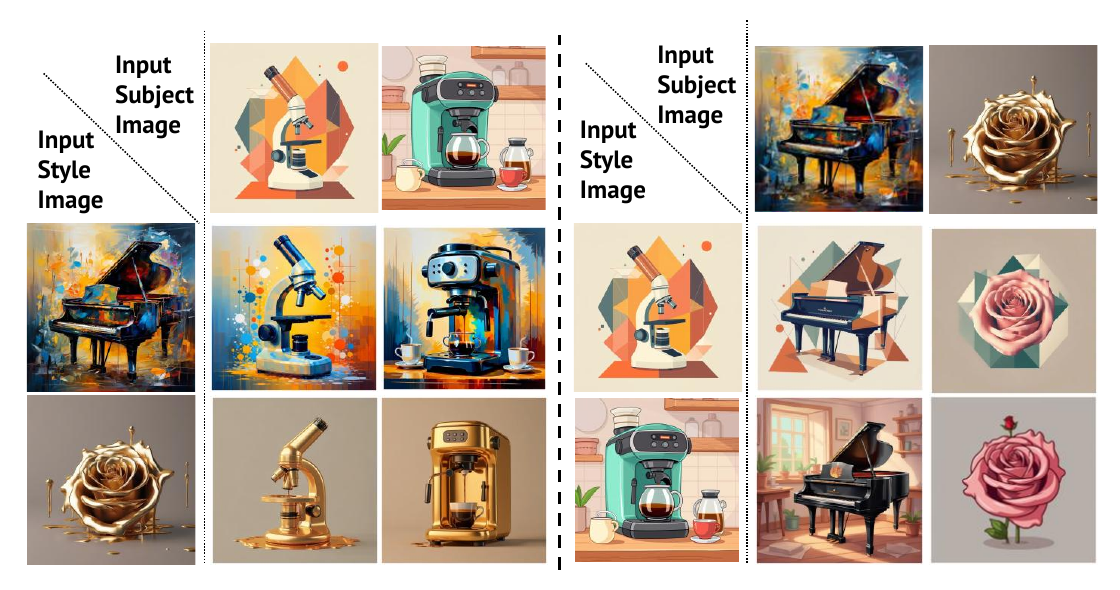}
    \caption{Additional examples of \textit{cross-composition} using subject and style from different input images. The result demonstrates that our method effectively integrates features from both subject and style.}
    \vspace{-0.8em}
    \label{fig:add_direct_combie_suppl}
\end{figure}

The subject and style LoRAs produced by UnZipLoRA open up a possibility for cross-combination: pairing a subject LoRA from one image with a style LoRA from another. Fig.~\ref{fig:add_direct_combie_suppl} provides additional results for such cross-combination where the LoRAs are combined by direct addition. While these LoRAs are not explicitly trained together (and thus not subject to the orthogonality constraints enforced by ZipLoRA~\cite{shah2024ziplora}), the inherent separation imposed by our column and block strategies generally results in higher compatibility than generic DreamBooth-LoRAs trained without such constraints. Consequently, direct arithmetic merger yields promising cross-stylization results.

\subsection{Additional Results on KOALA Diffusion}
Our method extends beyond SDXL, and is applicable to newer diffusion models. We demonstrate this generalizability by training on KOALA~\cite{Lee@koala}, a more efficient, recent text-to-image model with leaner architecture compared to SDXL. We provide additional results in Fig.~\ref{fig:add_kuala_suppl}. As shown, our method, when applied to KOALA, accurately captures subject and style and allows for successful recontextualization (though the overall quality of the results is not as high as for SDXL due to limited capacity and lower parameter count of KOALA).


\section{Compute Requirements}
UnZipLoRA achieves strong computational efficiency through a joint training strategy that optimizes resource utilization. By training both subject and style LoRAs concurrently in a single run, UnZipLoRA significantly reduces the overall training time.  Specifically, UnZipLoRA requires only $1260$ seconds to train both LoRAs on a single NVIDIA A40 GPU.

In contrast, most existing methods necessitate separate training processes for content and style, effectively doubling the time requirements. For instance, DreamBooth-LoRA requires $1860$ seconds per LoRA, resulting in a total training time of $3720$ seconds. While B-LoRA demonstrates faster individual LoRA training at $600$ seconds per LoRA ($1200$ seconds total), UnZipLoRA remains highly competitive. Notably, methods like Inspiration Tree incur significantly higher computational costs, requiring $7680$ seconds in total: $3840$s to select a good random seed for training and another $3840$s to train the model.

Beyond time efficiency, UnZipLoRA minimizes the number of parameters updated during training.  Through its block and column separation strategies, UnZipLoRA updates only up to $30\%$ of parameters in the downsampling block and bottleneck, and approximately $50\%$ in the upsampling block for each LoRA. This focused optimization reduces the trainable parameters by nearly $40\%$ compared to training two full LoRAs independently, further contributing to its efficiency. Owing to such efficient parameter utilization, UnZipLoRA exhibits faster convergence, requiring only $600$ steps of training as opposed to $800$ to $1000$ steps for most other methods including DreamBooth-LoRA and B-LoRA. 

\section{Image Attributions}
\label{sec:attr}
In our experiments, we use several stylized images as inputs images. We curate these input images from three sources: (i) free-to-use online repositories that provide artistic images; (ii) open-sourced repositories of previous works such as StyleDrop~\cite{sohn2023styledrop} and RB-Modulation~\cite{rout2024rbmodulation}; and (iii) synthetically generated images using freely available text-to-image models such as Flux. For the human-created artistic images, we provide image attributions below for each image that we used in our experiments.  

\noindent\textbf{Image attributions for the stylized images used as inputs} 

The sources of the style images that we used in our experiments are as follows:
\begin{itemize}
    \item \href{https://www.freepik.com/free-vector/tourist-hat-accessory_136482269.htm#fromView=search&page=1&position=44&uuid=c68c4e25-7b38-437f-bd78-1a9d90e46300}{Sun hat in flat cartoon illustration style}, 
\item \href{https://www.freepik.com/free-vector/hand-drawn-one-line-art-illustration_23164244.htm#fromView=search&page=1&position=30&uuid=b5562fcd-edea-44b7-8a57-7dff0c529c67}{Kangaroo in one line art illustration style}, 
\item \href{https://www.freepik.com/free-vector/smiley-kawaii-randoseru-with-stars_9925758.htm#fromView=search&page=1&position=42&uuid=768e5807-00d8-40fd-a450-0c8abf2a526b}{Backpack in cartoon illustration style}, 
\item \href{https://github.com/styledrop/styledrop.github.io/blob/main/images/assets/image_6487327_crayon_02.jpg}{A bear in kid crayon drawing style}, 
\item \href{https://github.com/google/RB-Modulation/blob/main/data/mosaic.png}{A teapot in mossaic art style}, 
\item \href{https://github.com/google/RB-Modulation/blob/main/data/linedrawing.png}{A telephone in line drawing style}
\end{itemize}

All the remaining input images are generated using Flux\footnote{\url{https://huggingface.co/black-forest-labs/FLUX.1-dev}} text-to-image diffusion model using the prompts provided by RB-modulation codebase on their github page at this URL: \href{https://github.com/google/RB-Modulation/blob/main/data/prompts.txt}{Text-prompts to generate stylized images}



%

\end{document}